# Exact Low Tubal Rank Tensor Recovery from Gaussian Measurements

Canyi Lu[1], Jiashi Feng[2], Zhouchen Lin[3,4] Shuicheng Yan[5,2]

[1] Department of Electrical and Computer Engineering, Carnegie Mellon University
[2] Department of Electrical and Computer Engineering, National University of Singapore
[3] Key Laboratory of Machine Perception (MOE), School of EECS, Peking University
[4] Cooperative Medianet Innovation Center, Shanghai Jiao Tong University
[5] 360 AI Institute
canyilu@gmail.com, elefjia@nus.edu.sg, zlin@pku.edu.cn, eleyans@nus.edu.sg

## Abstract

The recent proposed Tensor Nuclear Norm (TNN) [Lu *et al.*, 2016; 2018a] is an interesting convex penalty induced by the tensor SVD [Kilmer and Martin, 2011]. It plays a similar role as the matrix nuclear norm which is the convex surrogate of the matrix rank. Considering that the TNN based Tensor Robust PCA [Lu *et al.*, 2018a] is an elegant extension of Robust PCA with a similar tight recovery bound, it is natural to solve other low rank tensor recovery problems extended from the matrix cases. However, the extensions and proofs are generally tedious. The general atomic norm provides a unified view of low-complexity structures induced norms, e.g., the $\ell_1$-norm and nuclear norm. The sharp estimates of the required number of generic measurements for exact recovery based on the atomic norm are known in the literature. In this work, with a careful choice of the atomic set, we prove that TNN is a special atomic norm. Then by computing the Gaussian width of certain cone which is necessary for the sharp estimate, we achieve a simple bound for guaranteed low tubal rank tensor recovery from Gaussian measurements. Specifically, we show that by solving a TNN minimization problem, the underlying tensor of size $n_1 \times n_2 \times n_3$ with tubal rank $r$ can be exactly recovered when the given number of Gaussian measurements is $O(r(n_1 + n_2 - r)n_3)$. It is order optimal when comparing with the degrees of freedom $r(n_1 + n_2 - r)n_3$. Beyond the Gaussian mapping, we also give the recovery guarantee of tensor completion based on the uniform random mapping by TNN minimization. Numerical experiments verify our theoretical results.

## 1 Introduction

Many engineering problems look for solutions to underdetermined systems of linear equations: a system is considered underdetermined if there are fewer equations than unknowns. Suppose we are given information about an object $\mathbf{x}_0 \in \mathbb{R}^d$ of the form $\Phi \mathbf{x}_0 \in \mathbb{R}^m$ where $\Phi$ is an $m \times d$ matrix. We want the bound on the number of rows $m$ to ensure that $\mathbf{x}_0$ is the unique minimizer to the problem

$$\min_{\mathbf{x}} \|\mathbf{x}\|_A, \text{ s.t. } \Phi \mathbf{x}_0 = \Phi \mathbf{x}. \tag{1}$$

Here $\|\cdot\|_A$ is a norm with some suitable properties which encourage solutions to conform to some notion of simplicity. For example, the compressed sensing problem aims to recover a sparse vector $\mathbf{x}_0$ from (1) by taking $\|\cdot\|_A$ as the $\ell_1$-norm $\|\mathbf{x}\|_1$. We would like to know that how many measurements are required to recover an $s$-sparse $\mathbf{x}_0$. This of course depends on the kind of measurements. For instance, it is shown in [Candès *et al.*, 2006] that $20s \log d$ randomly selected Fourier coefficients are sufficient. If the Gaussian measurement map ($\Phi$ has entries i.i.d. sampled from a Gaussian distribution with mean 0 and variance $\frac{1}{m}$) is used, $2s \log \frac{d}{s} + \frac{5}{4}s$ measurements are needed [Donoho and Tanner, 2009; Chandrasekaran *et al.*, 2012]. Another interesting structured object is the low-rank matrix $\boldsymbol{X}_0 \in \mathbb{R}^{n_1 \times n_2}$. In this case, the $i$th component of a linear operator is given by $[\Phi(\boldsymbol{X}_0)]_i = \langle \boldsymbol{\Phi}_i, \boldsymbol{X}_0 \rangle$, where $\boldsymbol{\Phi}_i \in \mathbb{R}^{n_1 \times n_2}$. This includes the matrix completion problem [Candès and Recht, 2009] as a special case based on a proper choice of $\boldsymbol{\Phi}_i$. By taking $\|\cdot\|_A$ as the matrix nuclear norm $\|\boldsymbol{X}\|_*$, the convex program (1) recovers $\boldsymbol{X}_0$ provided that the number of measurements is of the order $\mu(\boldsymbol{X}_0) r(n_1 + n_2 - r) \log^2(n_1 + n_2)$, where $r$ is the rank of $\boldsymbol{X}_0$ and $\mu(\boldsymbol{X}_0)$ is the incoherence parameter [Candès and Recht, 2009; Chen, 2015]. Compared with the degrees of freedom $r(n_1 + n_2 - r)$ of the rank-$r$ matrix, such a rate is optimal (up to a logarithmic factor). If the Gaussian measurement map is used, about $3r(n_1 + n_2 - r)$ samples are sufficient for exact recovery [Recht *et al.*, 2010].

Beyond the sparse vector and low-rank matrix, there have some other structured signals which can be recovered by (1). The work [Chandrasekaran *et al.*, 2012] gives some more examples, presents a unified view of the convex programming to inverse problems and provides a simple framework to derive exact recovery bounds for a variety of simple models. Their considered models are formed as the sum of a few atoms from some elementary atomic sets. The convex programming formulation is based on minimizing the norm induced by the convex hull of the atomic set; this norm is referred to as the atomic norm (the $\ell_1$-norm and nuclear norm are special cases). By using the properties of the atomic norm, an analysis of the underlying convex geometry provides sharp estimates of the number of generic measurements required for exact recovery of models from partial information. A key step to estimate the required number of measurements is to compute the Gaussian width of the tangent cone associated with the atomic norm ball.

This work focuses the study on the low-rank tensor which is an interesting object structured that has many applications in signal processing. Recovering low-rank tensor is not easy since the tensor rank is not well defined. There have several tensor rank definitions, but each has its limitation. For example, the CP rank, defined as the smallest number of rank one tensor decomposition, is generally NP hard to compute. Also, its convex envelope is in general intractable. The tractable Tucker rank is more widely used. However, considering the low Tucker rank tensor recovery problem, the required number of measurements of existing convex model is much higher than the degrees of freedom [Mu *et al.*, 2014]. This is different from the nuclear norm minimization for low-rank matrix recovery which has order optimal rate [Chen, 2015].

In this work, we first study the low tubal rank tensor recovery from Gaussian measurements. Tensor RPCA [Lu *et al.*, 2016; 2018a] studies the low tubal rank tensor recovery from sparse corruptions by Tensor Nuclear Norm (TNN) minimization. We show that TNN is a new instance of the atomic norm based on a proper choice of the atomic set. From the perspective of atomic norm minimization, we give the low tubal rank recovery guarantee from Gaussian measurements. Specifically, to recover a tensor of size $n_1 \times n_2 \times n_3$ with tubal rank $r$ from Gaussian measurement by TNN minimization, the required number of measurements is $O(r(n_1 + n_2 - r)n_3)$. It is order optimal when comparing with the degrees of freedom $r(n_1 + n_2 - r)n_3$. Second, we study the tensor completion problem from uniform random sampling. We show that, to recover a tensor of tubal rank $r$, the sampling complexity is $O(r \min(n_1, n_2)n_3 \log^2(\min(n_1, n_2)n_3))$, which is order optimal (up to a log factor). The same problem has been studied in [Zhang and Aeron, 2017] but its proofs have several errors.

## 2 Notations and Preliminaries

We introduce some notations used in this paper. We denote tensors by boldface Euler script letters, e.g., $\mathcal{A}$, matrices by boldface capital letters, e.g., $\boldsymbol{A}$, vectors by boldface lowercase letters, e.g., $\boldsymbol{a}$, and scalars by lowercase letters, e.g., $a$. We denote $\boldsymbol{I}_n$ as the $n \times n$ sized identity matrix. The field of real number and complex number are denoted as $\mathbb{R}$ and $\mathbb{C}$, respectively. For a 3-way tensor $\mathcal{A} \in \mathbb{C}^{n_1 \times n_2 \times n_3}$, we denote its $(i, j, k)$-th entry as $\mathcal{A}_{ijk}$ or $a_{ijk}$ and use the Matlab notation $\mathcal{A}(i, :, :)$, $\mathcal{A}(:, i, :)$ and $\mathcal{A}(:, :, i)$ to respectively denote the $i$-th horizontal, lateral and frontal slice. More often, the frontal slice $\mathcal{A}(:, :, i)$ is denoted compactly as $\boldsymbol{A}^{(i)}$. The tube is denoted as $\mathcal{A}(i, j, :)$. The inner product of $\boldsymbol{A}$ and $\boldsymbol{B}$ in $\mathbb{C}^{n_1 \times n_2}$ is defined as $\langle \boldsymbol{A}, \boldsymbol{B} \rangle = \text{Tr}(\boldsymbol{A}^* \boldsymbol{B})$, where $\boldsymbol{A}^*$ denotes the conjugate transpose of $\boldsymbol{A}$ and $\text{Tr}(\cdot)$ denotes the matrix trace. The inner product of $\mathcal{A}$ and $\mathcal{B}$ in $\mathbb{C}^{n_1 \times n_2 \times n_3}$ is defined as $\langle \mathcal{A}, \mathcal{B} \rangle = \sum_{i=1}^{n_3} \langle \boldsymbol{A}^{(i)}, \boldsymbol{B}^{(i)} \rangle$. For any $\mathcal{A} \in \mathbb{C}^{n_1 \times n_2 \times n_3}$, the complex conjugate of $\mathcal{A}$ is denoted as $\text{conj}(\mathcal{A})$, which takes the complex conjugate of all entries of $\mathcal{A}$. We denote $\lfloor t \rfloor$ as the nearest integer less than or equal to $t$ and $\lceil t \rceil$ as the one greater than or equal to $t$. We denote the $\ell_1$-norm as $\|\mathcal{A}\|_1 = \sum_{ijk} |a_{ijk}|$, the infinity norm as $\|\mathcal{A}\|_\infty = \max_{ijk} |a_{ijk}|$ and the Frobenius norm as $\|\mathcal{A}\|_F = \sqrt{\sum_{ijk} |a_{ijk}|^2}$. The same norms are used for matrices and vectors. The spectral norm of a matrix $\boldsymbol{A}$ is denoted as $\|\boldsymbol{A}\| = \max_i \sigma_i(\boldsymbol{A})$, where $\sigma_i(\boldsymbol{A})$'s are the singular values of $\boldsymbol{A}$. The matrix nuclear norm is $\|\boldsymbol{A}\|_* = \sum_i \sigma_i(\boldsymbol{A})$.

For $\mathcal{A} \in \mathbb{R}^{n_1 \times n_2 \times n_3}$, by using the Matlab command fft, we denote $\bar{\mathcal{A}} \in \mathbb{C}^{n_1 \times n_2 \times n_3}$ as the result of Fast Fourier Transformation (FFT) of $\mathcal{A}$ along the 3-rd dimension, i.e., $\bar{\mathcal{A}} = \text{fft}(\mathcal{A}, [\,], 3)$. In the same fashion, we can compute $\mathcal{A}$ from $\bar{\mathcal{A}}$ using the inverse FFT, i.e., $\mathcal{A} = \text{ifft}(\bar{\mathcal{A}}, [\,], 3)$. In particular, we denote $\bar{\boldsymbol{A}}$ as a block diagonal matrix with $i$-th block on the diagonal as the frontal slice $\bar{\boldsymbol{A}}^{(i)}$ of $\bar{\mathcal{A}}$, i.e.,

$$\bar{\boldsymbol{A}} = \text{bdiag}(\bar{\mathcal{A}}) = \begin{bmatrix} \bar{\boldsymbol{A}}^{(1)} & & & \\ & \bar{\boldsymbol{A}}^{(2)} & & \\ & & \ddots & \\ & & & \bar{\boldsymbol{A}}^{(n_3)} \end{bmatrix}.$$

The block circulant matrix of $\mathcal{A}$ is defined as

$$\text{bcirc}(\mathcal{A}) = \begin{bmatrix} \boldsymbol{A}^{(1)} & \boldsymbol{A}^{(n_3)} & \cdots & \boldsymbol{A}^{(2)} \\ \boldsymbol{A}^{(2)} & \boldsymbol{A}^{(1)} & \cdots & \boldsymbol{A}^{(3)} \\ \vdots & \vdots & \ddots & \vdots \\ \boldsymbol{A}^{(n_3)} & \boldsymbol{A}^{(n_3-1)} & \cdots & \boldsymbol{A}^{(1)} \end{bmatrix}.$$

The block circulant matrix can be block diagonalized, i.e.,

$$(\boldsymbol{F}_{n_3} \otimes \boldsymbol{I}_{n_1}) \cdot \text{bcirc}(\mathcal{A}) \cdot (\boldsymbol{F}_{n_3}^{-1} \otimes \boldsymbol{I}_{n_2}) = \bar{\boldsymbol{A}},$$

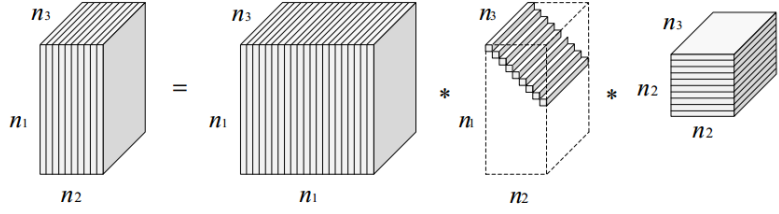

Figure 1: Illustration of the t-SVD of an $n_1 \times n_2 \times n_3$ tensor.

where $\boldsymbol{F}_{n_3} \in \mathbb{C}^{n_3 \times n_3}$ is the discrete Fourier transformation matrix, $\otimes$ denotes the Kronecker product. Note that $(\boldsymbol{F}_{n_3} \otimes \boldsymbol{I}_{n_1})/\sqrt{n_3}$ is orthogonal. We define the following operators

$$\texttt{unfold}(\boldsymbol{\mathcal{A}}) = \begin{bmatrix} \boldsymbol{A}^{(1)} \\ \boldsymbol{A}^{(2)} \\ \vdots \\ \boldsymbol{A}^{(n_3)} \end{bmatrix}, \ \texttt{fold}(\texttt{unfold}(\boldsymbol{\mathcal{A}})) = \boldsymbol{\mathcal{A}}.$$

**Definition 1.** *(t-product) [Kilmer and Martin, 2011] Let $\boldsymbol{\mathcal{A}} \in \mathbb{R}^{n_1 \times n_2 \times n_3}$ and $\boldsymbol{\mathcal{B}} \in \mathbb{R}^{n_2 \times l \times n_3}$. Then the t-product $\boldsymbol{\mathcal{A}} * \boldsymbol{\mathcal{B}}$ is defined to be a tensor $\boldsymbol{\mathcal{C}} \in \mathbb{R}^{n_1 \times l \times n_3}$,*

$$\boldsymbol{\mathcal{C}} = \boldsymbol{\mathcal{A}} * \boldsymbol{\mathcal{B}} = \texttt{fold}(\texttt{bcirc}(\boldsymbol{\mathcal{A}}) \cdot \texttt{unfold}(\boldsymbol{\mathcal{B}})).$$

The frontal slices of $\bar{\boldsymbol{\mathcal{A}}}$ has the following property

$$\begin{cases} \bar{\boldsymbol{A}}^{(1)} \in \mathbb{R}^{n_1 \times n_2}, \\ \texttt{conj}(\bar{\boldsymbol{A}}^{(i)}) = \bar{\boldsymbol{A}}^{(n_3-i+2)}, \ i = 2, \cdots, \lfloor \frac{n_3+1}{2} \rfloor. \end{cases} \quad (2)$$

Using the above property, the work [Lu *et al.*, 2018a] proposes a more efficient way for computing t-product than the method in [Kilmer and Martin, 2011].

**Definition 2.** *(Conjugate transpose) [Lu et al., 2016; 2018a] The conjugate transpose of a tensor $\boldsymbol{\mathcal{A}}$ of size $n_1 \times n_2 \times n_3$ is the $n_2 \times n_1 \times n_3$ tensor $\boldsymbol{\mathcal{A}}^*$ obtained by conjugate transposing each of the frontal slice and then reversing the order of transposed frontal slices 2 through $n_3$.*

**Definition 3.** *(Identity tensor) [Kilmer and Martin, 2011] The identity tensor $\boldsymbol{\mathcal{I}} \in \mathbb{R}^{n \times n \times n_3}$ is the tensor whose first frontal slice is the $n \times n$ identity matrix, and other frontal slices are all zeros.*

**Definition 4.** *(Orthogonal tensor) [Kilmer and Martin, 2011] A tensor $\boldsymbol{\mathcal{Q}} \in \mathbb{R}^{n \times n \times n_3}$ is orthogonal if it satisfies*

$$\boldsymbol{\mathcal{Q}}^* * \boldsymbol{\mathcal{Q}} = \boldsymbol{\mathcal{Q}} * \boldsymbol{\mathcal{Q}}^* = \boldsymbol{\mathcal{I}}.$$

**Definition 5.** *(F-diagonal Tensor) [Kilmer and Martin, 2011] A tensor is called f-diagonal if each of its frontal slices is a diagonal matrix.*

**Theorem 1.** *(T-SVD) [Lu et al., 2018a; Kilmer and Martin, 2011] Let $\boldsymbol{\mathcal{A}} \in \mathbb{R}^{n_1 \times n_2 \times n_3}$. Then it can be factored as*

$$\boldsymbol{\mathcal{A}} = \boldsymbol{\mathcal{U}} * \boldsymbol{\mathcal{S}} * \boldsymbol{\mathcal{V}}^*,$$

*where $\boldsymbol{\mathcal{U}} \in \mathbb{R}^{n_1 \times n_1 \times n_3}$, $\boldsymbol{\mathcal{V}} \in \mathbb{R}^{n_2 \times n_2 \times n_3}$ are orthogonal, and $\boldsymbol{\mathcal{S}} \in \mathbb{R}^{n_1 \times n_2 \times n_3}$ is a f-diagonal tensor.*

Theorem 1 gives the t-SVD based on t-product. See Figure 1 for an illustration. Theorem 1 appears first in [Kilmer and Martin, 2011] but their proof is not rigorous since it cannot guarantee that $\boldsymbol{\mathcal{U}}$ and $\boldsymbol{\mathcal{V}}$ are real tensors. The work [Lu *et al.*, 2018a] fixes this issue by using property (2), and further gives a more efficient way for computing t-SVD (see Algorithm 1). Algorithm 1 only needs to compute $\lceil \frac{n_3+1}{2} \rceil$ matrix SVDs, while this number is $n_3$ by the method in [Kilmer and Martin, 2011]. The entries of the first frontal slice $\boldsymbol{\mathcal{S}}(:,:,1)$ are called as the singular values of the tensor $\boldsymbol{\mathcal{A}}$. The number of nonzero singular values is equivalent to the tensor tubal rank.

**Definition 6.** *(Tensor tubal rank) [Lu et al., 2018a] For $\boldsymbol{\mathcal{A}} \in \mathbb{R}^{n_1 \times n_2 \times n_3}$, the tensor tubal rank, denoted as $\text{rank}_t(\boldsymbol{\mathcal{A}})$, is defined as the number of nonzero singular values of $c\boldsymbol{\mathcal{S}}$, where $\boldsymbol{\mathcal{S}}$ is from the t-SVD of $\boldsymbol{\mathcal{A}} = \boldsymbol{\mathcal{U}} * \boldsymbol{\mathcal{S}} * \boldsymbol{\mathcal{V}}^*$. We can write*

$$\text{rank}_t(\boldsymbol{\mathcal{A}}) = \#\{i, \boldsymbol{\mathcal{S}}(i,i,1) \neq 0\} = \#\{i, \boldsymbol{\mathcal{S}}(i,i,:) \neq 0\}.$$

For $\boldsymbol{\mathcal{A}} \in \mathbb{R}^{n_1 \times n_2 \times n_3}$ with tubal rank $r$, it has the skinny t-SVD, i.e., $\boldsymbol{\mathcal{A}} = \boldsymbol{\mathcal{U}} * \boldsymbol{\mathcal{S}} * \boldsymbol{\mathcal{V}}^*$, where $\boldsymbol{\mathcal{U}} \in \mathbb{R}^{n_1 \times r \times n_3}$, $\boldsymbol{\mathcal{S}} \in \mathbb{R}^{r \times r \times n_3}$, and $\boldsymbol{\mathcal{V}} \in \mathbb{R}^{n_2 \times r \times n_3}$, in which $\boldsymbol{\mathcal{U}}^* * \boldsymbol{\mathcal{U}} = \boldsymbol{\mathcal{I}}$ and $\boldsymbol{\mathcal{V}}^* * \boldsymbol{\mathcal{V}} = \boldsymbol{\mathcal{I}}$. We use the skinny t-SVD throughout this paper.

**Algorithm 1** T-SVD

**Input:** $\mathcal{A} \in \mathbb{R}^{n_1 \times n_2 \times n_3}$.
**Output:** T-SVD components $\mathcal{U}$, $\mathcal{S}$ and $\mathcal{V}$ of $\mathcal{A}$.

1. Compute $\bar{\mathcal{A}} = \texttt{fft}(\mathcal{A}, [\,], 3)$.
2. Compute each frontal slice of $\bar{\mathcal{U}}$, $\bar{\mathcal{S}}$ and $\bar{\mathcal{V}}$ from $\bar{\mathcal{A}}$ by
   **for** $i = 1, \cdots, \lceil \frac{n_3+1}{2} \rceil$ **do**
   $\quad [\bar{U}^{(i)}, \bar{S}^{(i)}, \bar{V}^{(i)}] = \text{SVD}(\bar{A}^{(i)})$;
   **end for**
   **for** $i = \lceil \frac{n_3+1}{2} \rceil + 1, \cdots, n_3$ **do**
   $\quad \bar{U}^{(i)} = \texttt{conj}(\bar{U}^{(n_3-i+2)})$;
   $\quad \bar{S}^{(i)} = \bar{S}^{(n_3-i+2)}$;
   $\quad \bar{V}^{(i)} = \texttt{conj}(\bar{V}^{(n_3-i+2)})$;
   **end for**
3. Compute $\mathcal{U} = \texttt{ifft}(\bar{\mathcal{U}}, [\,], 3)$, $\mathcal{S} = \texttt{ifft}(\bar{\mathcal{S}}, [\,], 3)$, and $\mathcal{V} = \texttt{ifft}(\bar{\mathcal{V}}, [\,], 3)$.

**Definition 7.** *(Tensor nuclear norm)* *[Lu* et al.*, 2018a] Let $\mathcal{A} = \mathcal{U} * \mathcal{S} * \mathcal{V}^*$ be the t-SVD of $\mathcal{A} \in \mathbb{R}^{n_1 \times n_2 \times n_3}$. The tensor nuclear norm of $\mathcal{A}$ is defined as the sum of the tensor singular values, i.e., $\|\mathcal{A}\|_* = \sum_{i=1}^{r} \mathcal{S}(i,i,1)$, where $r = rank_t(\mathcal{A})$.*

The above definition of TNN is defined based on t-SVD. It is equivalent to $\frac{1}{n_3} \|\bar{A}\|_*$ as given in [Lu *et al.*, 2016]. Indeed,

$$\|\mathcal{A}\|_* = \sum_{i=1}^{r} \mathcal{S}(i,i,1) = \langle \mathcal{S}, \mathcal{I} \rangle = \frac{1}{n_3} \langle \bar{\mathcal{S}}, \bar{\mathcal{I}} \rangle$$

$$= \frac{1}{n_3} \langle \bar{S}, \bar{I} \rangle = \frac{1}{n_3} \sum_{i=1}^{n_3} \|\bar{A}^{(i)}\|_* = \frac{1}{n_3} \|\bar{A}\|_*.$$

Above the factor $\frac{1}{n_3}$ is from the property $\|F_{n_3}\|_F^2 = n_3$, where $F_{n_3}$ is the discrete Fourier transformation matrix.

**Definition 8.** *(Tensor spectral norm)* *[Lu* et al.*, 2016] The tensor spectral norm of $\mathcal{A} \in \mathbb{R}^{n_1 \times n_2 \times n_3}$, denoted as $\|\mathcal{A}\|$, is defined as $\|\mathcal{A}\| = \|\texttt{bcirc}(\mathcal{A})\|$.*

TNN is the dual norm of the tensor spectral norm, and vice versa. Definite the tensor average rank as $\text{rank}_a(\mathcal{A}) = \frac{1}{n_3} \texttt{bcirc}(\mathcal{A})$. Then the convex envelope of the tensor average rank is the tensor nuclear within the set $\{\mathcal{A} | \|\mathcal{A}\| \leq 1\}$. It is worth mentioning that the above definition of tensor nuclear norm is different from the one in [Zhang and Aeron, 2017] due to the factor $\frac{1}{n_3}$. This factor is crucial in theoretical analysis. Intuitively, it makes the model, theoretical proof and the way for optimization consistent with the matrix cases.

## 3 Tensor Nuclear Norm Is an Atomic Norm

Based on the above tensor tubal rank, this work considers the following problem. Suppose that we have a linear map $\Phi : \mathbb{R}^{n_1 \times n_2 \times n_3} \to \mathbb{R}^m$ and the observations $\mathbf{y} = \Phi(\mathcal{M})$ for $\mathcal{M} \in \mathbb{R}^{n_1 \times n_2 \times n_3}$ which has tubal rank $r$. Our goal is to recover the underlying $\mathcal{M}$ from the observations $\mathbf{y}$. This can be achieved by solving the following convex program

$$\hat{\mathcal{X}} = \arg\min_{\mathcal{X}} \|\mathcal{X}\|_*, \text{ s.t. } \mathbf{y} = \Phi(\mathcal{X}). \tag{3}$$

Now, how many measurements are required to guarantee the *exact recovery* (i.e., $\hat{\mathcal{X}} = \mathcal{M}$)? This problem is an extension of the low-rank matrix recovery problem [Recht *et al.*, 2010]. To answer the above question, we will use the unified theory in [Chandrasekaran *et al.*, 2012] which provides sharp estimates of the number of measurements required for exact and robust recovery of models from Gaussian measurements. The key challenge is to reformulate TNN as a special case of the atomic norm and compute the Gaussian width. In this section, we will show that TNN is a special case of the atomic norm.

Let $A$ be a collection of atoms that is a compact subset of $\mathbb{R}^p$ and $\text{conv}(A)$ be its convex hull. The atomic norm induced by $A$ is defined as [Chandrasekaran *et al.*, 2012]

$$\|\mathbf{x}\|_A = \inf \left\{ \sum_{\mathbf{a} \in A} c_\mathbf{a} : \mathbf{x} = \sum_{\mathbf{a} \in A} c_\mathbf{a} \mathbf{a}, c_\mathbf{a} \geq 0, \forall \mathbf{a} \in A \right\}.$$

We also need some other notations which will be used in the analysis. The support function of $A$ is given as
$$\|\mathbf{x}\|_A^* = \sup\{\langle \mathbf{x}, \mathbf{a}\rangle : \mathbf{a} \in A\}.$$
If $\|\cdot\|_A$ is a norm, the support function $\|\cdot\|_A^*$ is the dual norm of this atomic norm.

A convex set $C$ is a cone if it is closed under positive linear combinations. The polar $C^*$ of a cone $C$ is the cone
$$C^* = \{\mathbf{x} \in \mathbb{R}^p : \langle \mathbf{x}, \mathbf{z}\rangle \leq 0, \forall \mathbf{z} \in C\}.$$
The tangent cone at nonzero $\mathbf{x}$ is defined as
$$T_A(\mathbf{x}) = \text{cone}\{\mathbf{z} - \mathbf{x} : \|\mathbf{z}\|_A \leq \|\mathbf{x}\|_A\}.$$
The normal cone $N_A(\mathbf{x})$ at $\mathbf{x}$ is defined as
$$N_A(\mathbf{x}) = \{\mathbf{s} : \langle \mathbf{s}, \mathbf{z} - \mathbf{s}\rangle \leq 0, \forall \mathbf{z} \text{ s.t. } \|\mathbf{z}\|_A \leq \|\mathbf{x}\|_A\}.$$
Note that the normal cone $N_A(\mathbf{x})$ is the conic hull of the subdifferential of the atomic norm at $\mathbf{x}$.

By a proper choice of the set $A$, the atomic norm reduces to several well-known norms. For example, let $A \subset \mathbb{R}^p$ be the set of unit-norm one-sparse vectors $\{\pm \mathbf{e}_i\}_{i=1}^p$. Then $k$-sparse vectors in $\mathbb{R}^p$ can be constructed using a linear combination of $k$ elements of the atomic set and the atomic norm $\|\mathbf{x}\|_A$ reduces to the $\ell_1$-norm. Let $A$ be the set of rank-one matrices of unit-Euclidean-norm. Then the rank-$k$ matrices can be constructed using a linear combination of $k$ elements of the atomic set and the atomic norm reduces to the matrix nuclear norm. Some other examples of atomic norms can be found in [Chandrasekaran et al., 2012]. At the following, we define a new atomic set $A$, and show that TNN is also an atomic norm induced by such an atomic set.

Let $D$ be a set of the following matrices, i.e., $\bar{\boldsymbol{D}} \in D$ where
$$\bar{\boldsymbol{D}} = \begin{bmatrix} \boldsymbol{D}_1 & & & \\ & \boldsymbol{D}_2 & & \\ & & \ddots & \\ & & & \boldsymbol{D}_{n_3} \end{bmatrix} \in \mathbb{C}^{n_1 n_3 \times n_2 n_3},$$
where $\boldsymbol{D}_i \in \mathbb{C}^{n_1 \times n_2}$ and there exists $k$ such that $\boldsymbol{D}_k \neq 0$, $\text{rank}(\boldsymbol{D}_k) = 1$, $\|\boldsymbol{D}_k\|_F = 1$, and $\boldsymbol{D}_j = 0$, for all $j \neq k$. Then, for any $\mathcal{A} \in \mathbb{R}^{n_1 \times n_2 \times n_3}$, we have
$$\|\bar{\boldsymbol{A}}\|_* = \inf\left\{\sum_{\bar{\boldsymbol{D}} \in D} c_{\bar{\boldsymbol{D}}} : \bar{\boldsymbol{A}} = \sum_{\bar{\boldsymbol{D}} \in D} c_{\bar{\boldsymbol{D}}} \bar{\boldsymbol{D}}, c_{\bar{\boldsymbol{D}}} \geq 0, \forall \bar{\boldsymbol{D}} \in D\right\}.$$
Above we use the property of the rank one matrix decomposition of a matrix. This is equivalent to
$$\|\bar{\boldsymbol{A}}\|_* = \inf\left\{\sum_{\bar{\mathcal{D}} \in D} c_{\bar{\mathcal{D}}} : \bar{\boldsymbol{A}} = \sum_{\bar{\mathcal{D}} \in D} c_{\bar{\mathcal{D}}} \bar{\mathcal{D}}, c_{\bar{\mathcal{D}}} \geq 0, \forall \bar{\mathcal{D}} \in D\right\}$$
$$= \inf\left\{\sum_{\bar{\mathcal{D}} \in D} c_{\mathcal{D}} : \mathcal{A} = \sum_{\bar{\mathcal{D}} \in D} c_{\mathcal{D}} \mathcal{D}, c_{\mathcal{D}} \geq 0, \forall \bar{\mathcal{D}} \in D\right\}, \quad (4)$$
where (4) uses the linear property of the inverse discrete Fourier transformation along the third dimension of a three way tensor. Motivated by (4), we define the atomic set $A$ as
$$A = \{\mathcal{W} \in \mathbb{C}^{n_1 \times n_2 \times n_3} : \mathcal{W} = n_3 \mathcal{D}, \bar{\mathcal{D}} \in D\}. \quad (5)$$
By $\|\mathcal{A}\|_* = \frac{1}{n_3}\|\bar{\boldsymbol{A}}\|_*$, we have the following result.

**Theorem 2.** *Let $A$ be the set defined as in (5). The atomic norm $\|\mathcal{A}\|_A$ is TNN, i.e.,*
$$\|\mathcal{A}\|_* = \|\mathcal{A}\|_A$$
$$= \inf\left\{\sum_{\mathcal{W} \in A} c_{\mathcal{W}} : \mathcal{A} = \sum_{\mathcal{W} \in A} c_{\mathcal{W}} \mathcal{W}, c_{\mathcal{W}} \geq 0, \forall \mathcal{W} \in A\right\}.$$

For any $\mathcal{W} \in A$, we have $\|\mathcal{W}\|_* = n_3 \|\mathcal{D}\|_* = \|\bar{\boldsymbol{D}}\|_* = 1$. So the convex hull $\text{conv}(A)$ is the TNN ball in which TNN is less than or equal to one. Interpreting TNN as a special atomic norm by choosing a proper atomic set is crucial for the low-rank tensor recovery guarantee.

# 4 Low-rank Tensor Recovery from Gaussian Measurements

The Corollary 3.3 in [Chandrasekaran *et al.*, 2012] shows that $\mathbf{x}_0$ is the unique solution to problem (1) with high probability provided $m \geq \omega^2(T_A(\mathbf{x}_0) \cap \mathbb{S}^{p-1}) + 1$. Here, $T_A(\mathbf{x}_0)$ is the tangent cone at $\mathbf{x}_0 \in \mathbb{R}^p$, $\mathbb{S}^{p-1}$ is the unit sphere, and $\omega(S)$ is the Gaussian width of a set $S$, defined as

$$\omega(S) = \mathbb{E}_{\mathbf{g}}\left[\sup_{\mathbf{z} \in S} \mathbf{g}^\top \mathbf{z}\right],$$

where $\mathbf{g}$ is a vector of independent zero-mean unit-variance Gaussians. To apply such a result for our low tubal rank recovery, we need to estimate the Gaussian width of our atomic set $A$ defined in (5).

**Theorem 3.** *Let $\mathcal{M} \in \mathbb{R}^{n_1 \times n_2 \times n_3}$ be a tubal rank $r$ tensor and $A$ in (5). We have that*

$$\omega(T_A(\mathcal{M}) \cap \mathbb{S}^{n_1 n_2 n_3 - 1}) \leq \sqrt{3r(n_1 + n_2 - r)n_3}. \tag{6}$$

Now, by using (6) and the Corollary 3.3 in [Chandrasekaran *et al.*, 2012], we have the following main result.

**Theorem 4.** *Let $\Phi : \mathbb{R}^{n_1 \times n_2 \times n_3} \to \mathbb{R}^n$ be a random map with i.i.d. zero-mean Gaussian entries having variance $\frac{1}{m}$ and $\mathcal{M} \in \mathbb{R}^{n_1 \times n_2 \times n_3}$ be a tensor of tubal rank $r$. Then, with high probability, we have:*

*(1) **exact recovery:** $\hat{\mathcal{X}} = \mathcal{M}$, where $\hat{\mathcal{X}}$ is the unique optimum of (3), provided that $m \geq 3r(n_1 + n_2 - r)n_3 + 1$;*

*(2) **robust recovery:** $\|\hat{\mathcal{X}} - \mathcal{M}\|_F \leq \frac{2\delta}{\epsilon}$, where $\hat{\mathcal{X}}$ is optimal to*

$$\hat{\mathcal{X}} = \arg\min_{\mathcal{X}} \|\mathcal{X}\|_*, \text{ s.t. } \|\mathbf{y} - \Phi(\mathcal{X})\|_2 \leq \delta, \tag{7}$$

*provided that $m \geq \frac{3r(n_1+n_2-r)n_3 + 3/2}{(1-\epsilon)^2}$.*

The above theorem shows that the tensor with tubal rank $r$ can be recovered exactly by solving the convex program (3) or approximately by (7) when the required number of measurements is of the order $O(r(n_1 + n_2 - r)n_3)$. Note that such a rate is optimal compared with the degrees of freedom of a tensor with tubal rank $r$.

**Theorem 5.** *A $n_1 \times n_2 \times n_3$ sized tensor with tubal rank $r$ has at most $r(n_1 + n_2 - r)n_3$ degrees of freedom.*

It is worth mentioning that the guarantee for low tubal rank tehsor recovery in Theorem 4 is an extension of the low matrix guarantee in [Recht *et al.*, 2010; Chandrasekaran *et al.*, 2012]. If $n_3 = 1$, the tensor $\mathcal{X}$ reduces to a matrix, the tensor tubal rank reduces to the matrix rank, and TNN reduces to the matrix nuclear norm. Thus the convex program (3) and the theoretical guarantee in Theorem 4 include the low rank matrix recovery model and guarantee as special cases, respectively. Compared with the existing low rank tensor recovery guarantees (based on different tensor ranks, e.g., [Mu *et al.*, 2014]) which are not order optimal, our guarantee enjoys the same optimal rate as the matrix case and our model (3) is computable.

# 5 Exact Tensor Completion Guarantee

Theorem 4 gives the recovery guarantee of program (3) based on the Gaussian measurements. In this section, we consider the tensor completion problem which is a special case of (3) but based on the uniform random mapping. Suppose that $\mathcal{M} \in \mathbb{R}^{n_1 \times n_2 \times n_3}$ and $\text{rank}_t(\mathcal{M}) = r$. We consider the Bernoulli model in this work: the entries of $\mathcal{M}$ are independently observed with probability $p$. We denote the set of the indices of the observed entries as $\Omega$. We simply denote $\Omega \sim \text{Ber}(p)$. Then, the tensor completion problem asks for recovering $\mathcal{M}$ from the observations $\{\mathcal{M}_{ij}, (i, j, k) \in \Omega\}$. We can solve this problem by solving the following program

$$\min_{\mathcal{X}} \|\mathcal{X}\|_*, \text{ s.t. } \mathcal{P}_\Omega(\mathcal{X}) = \mathcal{P}_\Omega(\mathcal{M}), \tag{8}$$

where $\mathcal{P}_\Omega(\mathcal{X})$ denotes the projection of $\mathcal{X}$ on the observed set $\Omega$. The above model extends the matrix completion task by convex nuclear norm minimization [Candès and Recht, 2009]. To guarantee the exact recovery, we need the following tensor incoherence conditions [Lu *et al.*, 2018a]

$$\max_{i=1,\cdots,n_1} \|\mathcal{U}^* * \mathring{\mathbf{e}}_i\|_F \leq \sqrt{\frac{\mu r}{n_1 n_3}}, \tag{9}$$

$$\max_{j=1,\cdots,n_2} \|\mathcal{V}^* * \mathring{\mathbf{e}}_j\|_F \leq \sqrt{\frac{\mu r}{n_2 n_3}}, \tag{10}$$

where $\mathring{\mathbf{e}}_i$ denotes the tensor column basis, which is a tensor of size $n \times 1 \times n_3$ with its $(i, 1, 1)$-th entry equaling 1 and the rest equaling 0. We also define the tensor tube basis $\dot{\mathbf{e}}_k$, which is a tensor of size $1 \times 1 \times n_3$ with its $(1, 1, k)$-th entry equaling 1 and the rest equaling 0. Denote $n_{(1)} = \max(n_1, n_2)$ and $n_{(2)} = \min(n_1, n_2)$.

Table 1: Exact low tubal rank tensor recovery from Gaussian measurements with sufficient number of measurements.

| $r = \text{rank}_t(\mathcal{X}_0) = 0.2n$ | | | | |
|---|---|---|---|---|
| $n$ | $\text{rank}_t(\mathcal{X}_0)$ | $m$ | $\text{rank}_t(\hat{\mathcal{X}})$ | $\frac{\|\hat{\mathcal{X}} - \mathcal{X}_0\|_F}{\|\mathcal{X}_0\|_F}$ |
| 10 | 2 | 541 | 2 | 1.2e−9 |
| 20 | 4 | 2161 | 4 | 1.6e−9 |
| 30 | 6 | 4861 | 6 | 1.5e−9 |
| $r = \text{rank}_t(\mathcal{X}_0) = 0.3n$ | | | | |
| $n$ | $\text{rank}_t(\mathcal{X}_0)$ | $m$ | $\text{rank}_t(\hat{\mathcal{X}})$ | $\frac{\|\hat{\mathcal{X}} - \mathcal{X}_0\|_F}{\|\mathcal{X}_0\|_F}$ |
| 10 | 3 | 766 | 3 | 1.6e−9 |
| 20 | 6 | 3061 | 6 | 1.2e−9 |
| 30 | 9 | 6886 | 9 | 1.2e−9 |

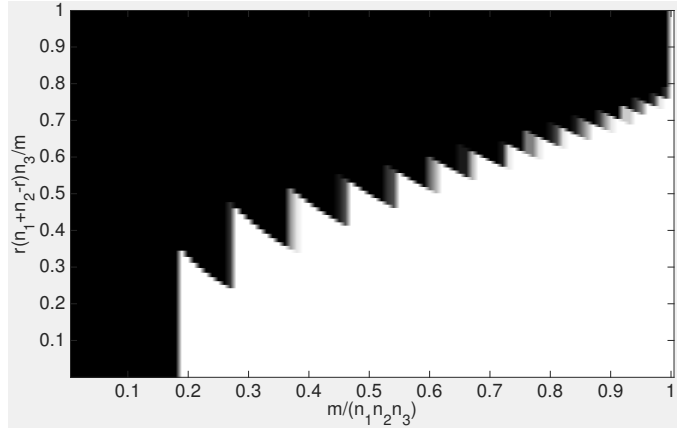

Figure 2: Phase transitions for low tubal rank tensor recovery from Gaussian measurements. Fraction of correct recoveries is across 10 trials, as a function of $\frac{r(n_1+n_2-r)n_3}{m}$ (y-axis) and sampling rate $\frac{m}{n_1 n_2 n_3}$. In this test, $n_1 = n_2 = 30, n_3 = 5$.

**Theorem 6.** *Let $\mathcal{M} \in \mathbb{R}^{n_1 \times n_2 \times n_3}$ with $\text{rank}_t(\mathcal{M}) = r$ and the skinny t-SVD be $\mathcal{M} = \mathcal{U} * \mathcal{S} * \mathcal{V}^*$. Suppose that the indices $\Omega \sim \text{Ber}(p)$ and the tensor incoherence conditions (9)-(10) hold. There exist universal constants $c_0, c_1, c_2 > 0$ such that if*

$$p \geq \frac{c_0 \mu r \log^2(n_{(1)} n_3)}{n_{(2)} n_3},$$

*then $\mathcal{M}$ is the unique solution to (8) with probability at least $1 - c_1(n_1 + n_2)^{-c_2}$.*

Theorem 6 shows that, to recover a $n_1 \times n_2 \times n_3$ sized tensor with tubal rank $r$, the sampling complexity is $O(rn_{(1)} n_3 \log^2(n_{(1)} n_3))$. Such a bound is tight compared with the degrees of freedom[1].

## 6 Experiments

In this section, we conducts experiments to first verify the exact recovery guarantee in Theorem 4 for (3) from Gaussian measurements, then to verify the exact recovery guarantee in Theorem 6 for tensor completion (8). Both (3) and (8) can be solved by the standard ADMM [Lu *et al.*, 2018b][2].

### 6.1 Exact Recovery from Gaussian Measurements

To verify Theorem 4, we can reformulate (3) as

$$\hat{\mathcal{X}} = \arg\min_{\mathcal{X}} \|\mathcal{X}\|_*, \text{ s.t. } \mathbf{y} = \mathbf{A}\text{vec}(\mathcal{X}), \tag{11}$$

where $\mathcal{X} \in \mathbb{R}^{n_1 \times n_2 \times n_3}$, $\mathbf{A} \in \mathbb{R}^{m \times (n_1 n_2 n_3)}$, $\mathbf{y} \in \mathbb{R}^m$ and $\text{vec}(\mathcal{X})$ denotes the vectorization of $\mathcal{X}$. The elements of $\mathbf{A}$ are with i.i.d. zero-mean Gaussian entries having variance $1/m$. Thus, $\mathbf{A}\text{vec}(\mathcal{X})$ gives the linear map $\Phi(\mathcal{X})$.

---

[1] The proofs in [Zhang and Aeron, 2017] for tensor completion have several errors. Their used TNN definition is different from ours.

[2] The codes of our methods can be found at https://github.com/canyilu/tensor-completion-tensor-recovery.

Table 2: Exact tensor completion on random data.
$\mathcal{X}_0 \in \mathbb{R}^{n \times n \times n}, r = \text{rank}_t(\mathcal{X}_0), m = pn^3, d_r = r(2n-r)n$

| $n$ | $r$ | $\frac{m}{d_r}$ | $p$ | $\text{rank}_t(\hat{\mathcal{X}})$ | $\frac{\|\hat{\mathcal{X}}-\mathcal{X}\|_F}{\|\mathcal{X}\|_F}$ |
|---|---|---|---|---|---|
| 50 | 3 | 4 | 0.47 | 3 | 3.9e−7 |
| 50 | 5 | 3 | 0.57 | 5 | 3.5e−7 |
| 50 | 10 | 2 | 0.72 | 10 | 4.1e−7 |
| 100 | 5 | 4 | 0.39 | 5 | 1.4e−6 |
| 100 | 10 | 3 | 0.57 | 10 | 9.2e−7 |
| 100 | 15 | 2 | 0.56 | 15 | 8.4e−7 |
| 200 | 5 | 4 | 0.20 | 5 | 4.2e−6 |
| 200 | 10 | 3 | 0.29 | 10 | 3.2e−6 |
| 200 | 20 | 2 | 0.38 | 20 | 3.1e−6 |
| 300 | 10 | 4 | 0.26 | 10 | 5.1e−6 |
| 300 | 20 | 3 | 0.39 | 20 | 4.2e−6 |
| 300 | 30 | 3 | 0.57 | 30 | 2.9e−6 |

First, we test on random tensors, provided sufficient number of measurements as suggested in Theorem 4. We generate $\mathcal{X}_0 \in \mathbb{R}^{n \times n \times n_3}$ of tubal rank $r$ by $\mathcal{X}_0 = \mathcal{P} * \mathcal{Q}$, where $\mathcal{P} \in \mathbb{R}^{n \times r \times n_3}$ and $\mathcal{Q} \in \mathbb{R}^{r \times n \times n_3}$ are with i.i.d. standard Gaussian random variables. We generate $\boldsymbol{A} \in \mathbb{R}^{m \times (n^2 n_3)}$ with its entries being i.i.d., zero-mean, $\frac{1}{m}$-variance Gaussian variables. Then, let $\mathbf{y} = \boldsymbol{A}\text{vec}(\mathcal{X}_0)$. We choose $n = 10, 20, 30$, $n_3 = 5$, $r = 0.2n$ and $r = 0.3n$. We set the number of measurements $m = 3r(2n-r)n_3 + 1$ as in Theorem 4. The results are given in Table 1, in which $\hat{\mathcal{X}}$ is the solution to (11). It can be seen that the relative errors $\|\hat{\mathcal{X}} - \mathcal{X}_0\|_F / \|\mathcal{X}_0\|_F$ are very small and the tubal ranks of $\hat{\mathcal{X}}$ are correct. Thus, this experiment verifies Theorem 4 for low tubal rank tensor recovery from Gaussian measurements.

Second, we exam the phase transition phenomenon in tubal rank $r$ and the number of measurements $m$. We set $n_1 = n_2 = 30$ and $n_3 = 5$. We vary $m$ between 1 and $n_1 n_2 n_3$ where the tensor is completely discovered. For a fixed $m$, we generate all possible tubal ranks such that $r(n_1 + n_2 - r)n_3 \leq m$. For each $(m, r)$ pair, we repeat the following procedure 10 times. We generate $\mathcal{X}_0$, $\boldsymbol{A}$, $\mathbf{y}$ in the same way as the first experiment above. We declare $\mathcal{X}_0$ to be recovered if $\|\hat{\mathcal{X}} - \mathcal{X}_0\|_F / \|\mathcal{X}_0\|_F \leq 10^{-3}$. Figure 2 plots the fraction of correct recovery for each pair. The color of the cell in the figure reflects the empirical recovery rate of the 10 runs (scaled between 0 and 1). In all experiments, white denotes perfect recovery, while black denotes failure. It can be seen that there is a large region in which the recovery is correct. When the underlying tubal rank $r$ of $\mathcal{X}_0$ is relatively larger, the required number of measurements for correct recovery is also larger. Such a result is consistent with our theoretical result. Similar phenomenon can be found in low-rank matrix recovery [Chandrasekaran et al., 2012].

### 6.2 Exact Tensor Completion

First, we verify the exact tensor completion guarantee in Theorem 6 on random data. We generate $\mathcal{M} \in \mathbb{R}^{n \times n \times n}$ with tubal rank $r$ by $\mathcal{M} = \mathcal{P} * \mathcal{Q}$, where the entries of $\mathcal{P} \in \mathbb{R}^{n \times r \times n}$ and $\mathcal{Q} \in \mathbb{R}^{r \times n \times n}$ are independently sampled from an $\mathcal{N}(0, 1/n)$ distribution. Then we sample $m = pn^3$ elements uniformly from $\mathcal{M}$ to form the known samples. A useful quantity for reference is the number of degrees of freedom $d_r = r(2n-r)n$. The results in Table 1 shows that program (8) gives the correct recovery in the sense that the relative errors are small, less than $10^{-5}$ and the tubal ranks of the obtained solution are correct. These results well verify the recovery guarantee in Theorem 6.

Second, we examine the recovery phenomenon with varying tubal rank of $\mathcal{M}$ and varying sampling rate $p$. We consider two sizes of $\mathcal{M} \in \mathbb{R}^{n \times n \times n}$: (1) $n = 40$; (2) $n = 50$. We generate $\mathcal{M} = \mathcal{P} * \mathcal{Q}$, where the entries of $\mathcal{P} \in \mathbb{R}^{n \times r \times n}$ and $\mathcal{Q} \in \mathbb{R}^{r \times n \times n}$ are independently sampled from an $\mathcal{N}(0, 1/n)$ distribution. We set $m = pn^3$. We choose $p$ in $[0.01 : 0.01 : 0.99]$ and $r = 1, 2, \ldots, 30$ in the case $n = 40$, and $r = 1, 2, \ldots, 35$ in the case $n = 50$. For each $(r, p)$ triple, we simulate 10 test instances and declare a trial to be successful if the recovered $\hat{\mathcal{X}}$ satisfies $\|\hat{\mathcal{X}} - \mathcal{M}\|_F / \|\mathcal{M}\|_F \leq 10^{-3}$. Figure 3 plots the fraction of correct recovery for each triple (black = 0% and white = 100%). It can be seen that there is a large region in which the recovery is correct. Interestingly, the experiments reveal very similar plots for different $n$, suggesting that our asymptotic conditions for recovery may be conservative. Such a phenomenon is also consistent with the result in Theorem 6 which shows that the recovery is correct when the sampling rate $p$ is not small and the tubal rank $r$ is relatively low.

## 7 Conclusion

This paper first considers the exact guarantee of TNN minimization for low tubal rank tensor recovery from Gaussian measurements. We prove that TNN is a new instance of the atomic norm associated with certain atomic set. From the perspective of atomic norm minimization, we give the optimal estimation of the required measurements for the exact low tubal rank tensor recovery. Second, we give the exact recovery guarantee of TNN minimization for tensor completion. This result fixes the errors in the proofs of [Zhang and Aeron, 2017]. Numerical experiments verify our theoretical results.

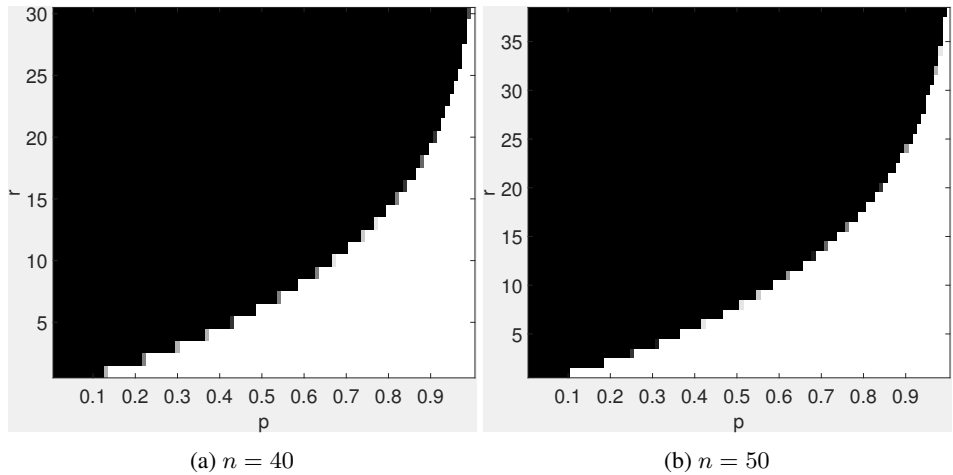

(a) $n = 40$     (b) $n = 50$

Figure 3: Phase transitions for tensor completion. Fraction of correct recoveries is across 10 trials, as a function of tubal rank $r$ (y-axis) and sampling rate $p$ (x-axis). The results are shown for different sizes of $\mathcal{M} \in \mathbb{R}^{n \times n \times n}$: (a) $n = 40$; (b) $n = 50$.

By treating TNN as an instance of the atomic norm, we can get more results of low tubal rank recovery by using existing results, e.g., [Foygel and Mackey, 2014; Amelunxen *et al.*, 2014]. Beyond the study on the convex TNN, it is also interesting to study the noncnovex models [Lu *et al.*, 2015].

## Acknowledgements

J. Feng is partially supported by National University of Singapore startup grant R-263-000-C08-133 and Ministry of Education of Singapore AcRF Tier One grant R-263-000-C21-112. Z. Lin was supported by National Basic Research Program of China (973 Program) (grant no. 2015CB352502), National Natural Science Foundation (NSF) of China (grant nos. 61625301 and 61731018), Qualcomm, and Microsoft Research Asia.

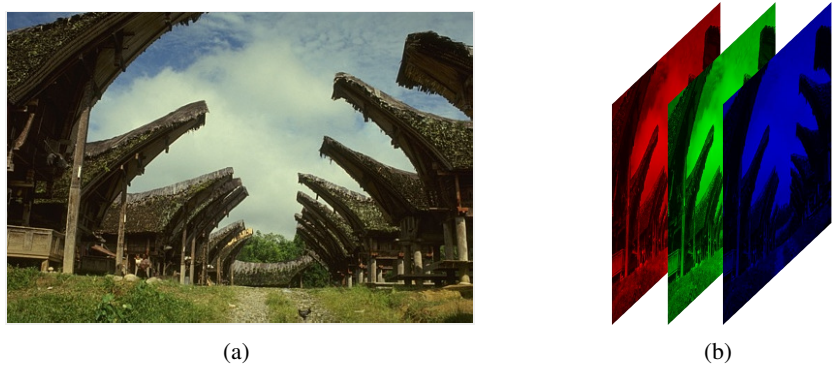

(a)              (b)

Figure 4: A color image of size $h \times w$ can be formated as a 3-way tensor $\mathcal{M} \in \mathbb{R}^{h \times 3 \times w}$, where the lateral slices correspond to the R, G, B channels.

## Appendix

## A  Applications of Tensor Completion on Real Data

### A.1  Tensor Completion for Image Inpainting

A color image has 3 channels, and thus it is a 3-way tensor in nature. It is observed that each channel can be approximated by low-rank matrix [Lu *et al.*, 2014]. Thus the matrix completion can be applied to recover the missing information of images, which may be corrupted by some noises, e.g., logos. However, applying matrix completion on each channel independently may degrade the performance. We consider tensor completion for color image recovery in this section.

For any color image of size $h \times w$, it can be formated as a 3-way tensor $\mathcal{M} \in \mathbb{R}^{h \times 3 \times w}$, where the lateral slices correspond to the three channels of the color image[3]. See Figure 4 for an illustration. We use this way of tensor construction from images as the input of tensor completion model in (8). We randomly select 100 color images from the Berkeley Segmentation Dataset [Martin *et al.*, 2001] for this test. We randomly set $m = 3phw$ entries to be observed. We consider $p = 0.3$ and $p = 0.5$ in this experiment. See Figure 5 (b) for some example images with missing values. Then we apply the following four methods for image recovery and compare their performance:

- LRMC: apply the low-rank matrix completion method [Candès and Recht, 2009] on each channel of images separably and combine the results.
- LRTC: low-rank tensor completion method in [Liu *et al.*, 2013]. We set the parameter $[\lambda_1 \ \lambda_2 \ \lambda_3] = \boldsymbol{\alpha}/\|\boldsymbol{\alpha}\|_1$, where $\boldsymbol{\alpha} = [1 \ 1 \ 10^{-3}]$.
- TMac: **t**ensor completion by parallel **ma**trix fa**c**torization method in [Xu *et al.*, 2015].
- TNN: apply the TNN based tensor completion model in (8) on the above way of tensor construction from images.

---

[3] We observe that the TNN based tensor completion model in (8) performs best in this way of tensor construction from a color image in most cases, though there have some other ways of tensor construction, e.g., $\mathcal{M} \in \mathbb{R}^{h \times w \times 3}$ as in [Zhang *et al.*, 2014].

We use the Peak Signal-to-Noise Ratio (PSNR), defined as

$$\text{PSNR} = 10 \log_{10} \left( \frac{\|\boldsymbol{\mathcal{M}}\|_\infty^2}{\frac{1}{n_1 n_2 n_3} \|\hat{\boldsymbol{\mathcal{X}}} - \boldsymbol{\mathcal{M}}\|_F^2} \right), \tag{12}$$

to evaluate the recovery performance. The higher PSNR value implies better recovery performance. Figure 6 and 7 show the PSNR values of the compared methods on all 100 images with the rate of observed entries $p = 0.3$ and $p = 0.5$, respectively. Some examples with the recovered images are shown in Figure 5. From these results, we have the following observations:

- The tensor based methods, including LRTC, TMac and TNN, usually perform much better than the matrix completion method LRMC. The reason is that LRMC, which performs the matrix completion on each channel independently, is not able to use the information across channels, while the tensor methods improve the performance by taking the advantage of the multi-dimensional structure of data. Such a phenomenon has also been observed in previous work [Liu *et al.*, 2013; Xu *et al.*, 2015].
- TNN based tensor completion model achieves better recovery performance than LRTC and TMac. This not only demonstrates the superiority of TNN, but also validates our recovery guarantee in Theorem 6 on image data. Both LRTC and TMac are sum of nuclear norm based methods and one needs some additional effort to tune the weighted parameters $\lambda_i$'s empirically. The obtained solution by LRTC is optimal, but it does not guarantee the lowest rank properties of the unfolded matrices of the tensor along different dimensions, since the sum of nuclear norm is a loose convex surrogate of the sum of rank. TMac solves the sum of nuclear norm based model more efficiently by matrix factorization, but it requires estimating the underlying ranks of the unfolded matrices. This is generally difficult without priori knowledge. There is no recovery guarantee of TMac either. In contrast, similar to the matrix nuclear norm, TNN is a tight convex relaxation of the tensor average rank, and the recovery performance of the obtained optimal solutions has the theoretical guarantee.

### A.2 Tensor Completion for Video Recovery

A grayscale video is a 3-way tensor in nature. In this section, we consider the video recovery problem by low-rank tensor completion from partially observed entries. We use 15 videos from http://trace.eas.asu.edu/yuv/ for the test. See Table 3 for all these 15 video sequences. For each sequence, we use the first 150 frames for the test due to the computational limitation. Note that the given videos are color videos. We convert them into grayscale, and thus they can be formated as 3-way tensors. For the sequences in Table 3, we use the file in the provided QCIF format, in which each frame has the size $144 \times 176$.

For a video with $f$ sequences and each frame has size $h \times w$, we can construct a tensor $\boldsymbol{\mathcal{M}} \in \mathbb{R}^{h \times f \times w}$. See Figure 8 for an illustration. We observe that the TNN based tensor completion model in (8) performs best in this way of tensor construction from videos in most cases. For a tensor $\boldsymbol{\mathcal{M}} \in \mathbb{R}^{h \times f \times w}$ constructed from a video, we randomly set $m = phfw$ entries to be observed, where we set $p = 0.5$ in this experiment. See Figure 9 (b) for some example frames with missing values. Then we apply LRMC, LRTC, TMac and TNN to complete $\mathcal{P}_\Omega(\boldsymbol{\mathcal{M}})$. In LRTC, we set $[\lambda_1 \ \lambda_2 \ \lambda_3] = [\frac{1}{3} \ \frac{1}{3} \ \frac{1}{3}]$. We evaluate the performance by using the PSNR values in (12). Table 3 shows the PSNR values of the compared methods on all 15 video sequences and the recovery results of some frames can be found in Figure 9. From these results, we can see that the TNN based tensor completion model in (8) also achieves best performance performance.

## B Optimization by ADMM

In this section, we give the optimization details for solving problems (11) and (8) by the standard ADMM [Lu *et al.*, 2018b].

First, problem (11) can be equivalently reformulated as

$$\min_{\boldsymbol{\mathcal{X}}, \boldsymbol{\mathcal{Z}}} \|\boldsymbol{\mathcal{X}}\|_*, \text{ s.t. } \mathbf{y} = \boldsymbol{A}\text{vec}(\boldsymbol{\mathcal{Z}}), \ \boldsymbol{\mathcal{X}} = \boldsymbol{\mathcal{Z}}. \tag{13}$$

The augmented Lagrangian function is

$$\mathcal{L}(\boldsymbol{\mathcal{X}}, \boldsymbol{\mathcal{Z}}, \boldsymbol{\lambda}_1, \boldsymbol{\lambda}_2) = \|\boldsymbol{\mathcal{X}}\|_* + \langle \boldsymbol{\lambda}_1, \boldsymbol{A}\text{vec}(\boldsymbol{\mathcal{Z}}) - \mathbf{y} \rangle + \langle \boldsymbol{\lambda}_2, \boldsymbol{\mathcal{X}} - \boldsymbol{\mathcal{Z}} \rangle + \frac{\mu}{2} \|\boldsymbol{A}\text{vec}(\boldsymbol{\mathcal{Z}}) - \mathbf{y}\|_F^2 + \frac{\mu}{2} \|\boldsymbol{\mathcal{X}} - \boldsymbol{\mathcal{Z}}\|_F^2,$$

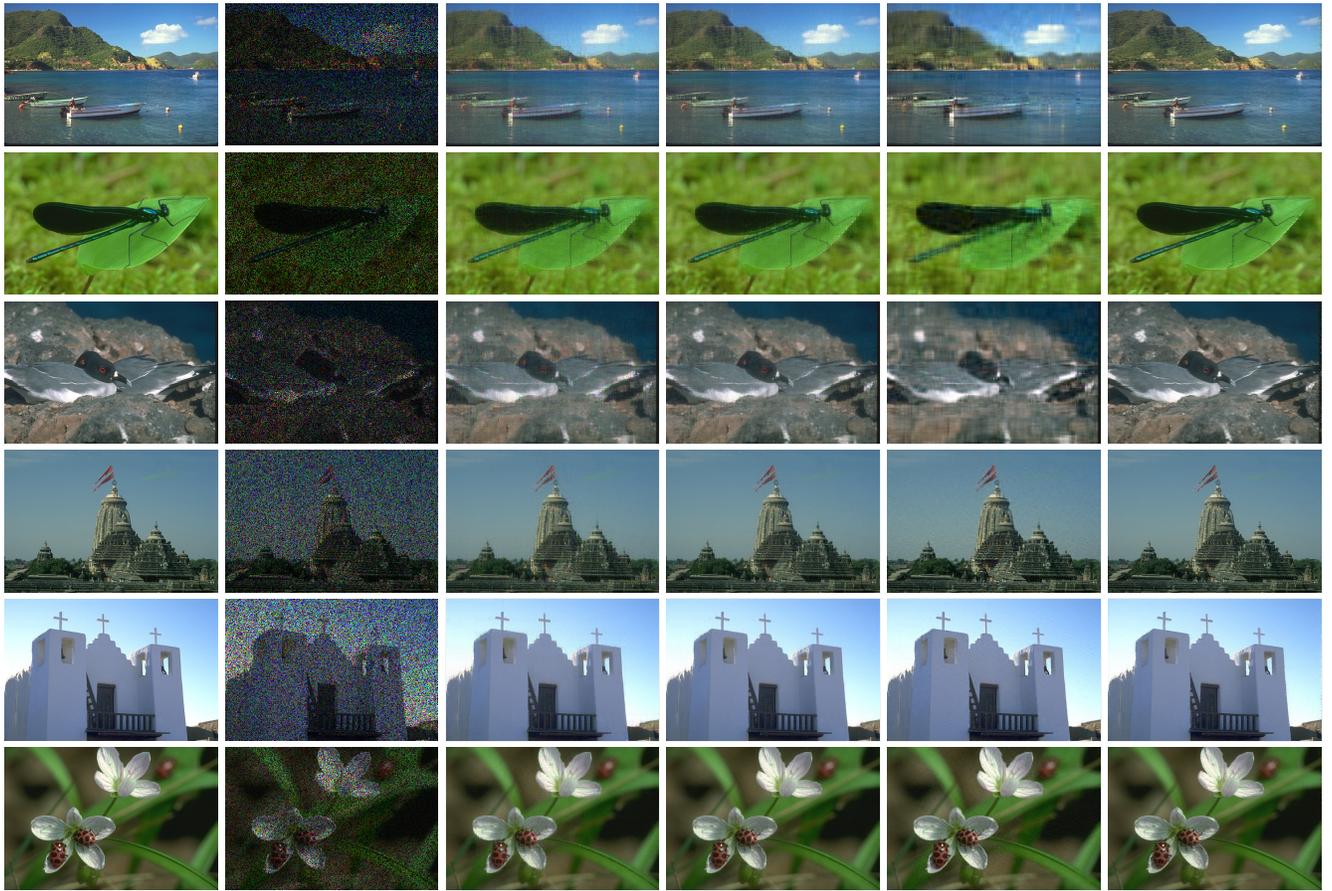

|  LRMC | LRTC | TMac | TNN  |
| --- | --- | --- | --- |
| 24.4 | 26.3 | 24.6 | **29.1** |
| 28.4 | 29.5 | 27.4 | **33.9** |
| 25.2 | 28.2 | 25.3 | **32.0** |

|  LRMC | LRTC | TMac | TNN  |
| --- | --- | --- | --- |
| 27.8 | 30.8 | 29.5 | **34.7** |
| 33.0 | 36.5 | 36.0 | **35.9** |
| 31.3 | 34.7 | 33.9 | **37.7** |

(g) PSNR values of the first three images with $p = 0.3$

(h) PSNR values of the last three images with $p = 0.5$

Figure 5: Examples for image recovery performance comparison. The first three rows are the results with $p = 0.3$ and the last three rows are the results with $p = 0.5$. (a) Original image; (b) observed image; (c)-(f) recovered images by LRMC, LRTC, TMac and TNN, respectively; (h) and (i) show the PSNR values obtained by the compared methods corresponding to the first thee rows and the last three rows, respectively.

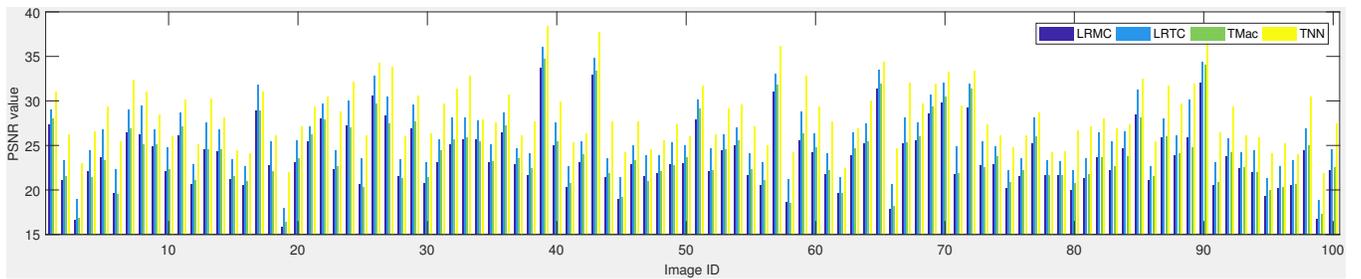

Figure 6: Comparison of the PSNR values obtained by using LRMC, LRTC, TMac and TNN. The rate of observed entries is $p = 0.3$.

where $\boldsymbol{\lambda}_1$ and $\boldsymbol{\lambda}_2$ are the dual variables. Then $\mathcal{X}$ and $\mathcal{Z}$ can be updated alternately by minimizing the augmented Lagrangian function. We show the updating details in Algorithm 2. Note that both updates of $\mathcal{X}$ and $\mathcal{Z}$ have closed form solutions. The update of $\mathcal{X}$ requires computing the proximal operator of TNN. Its closed form solution can be found at [Lu *et al.*, 2018a].

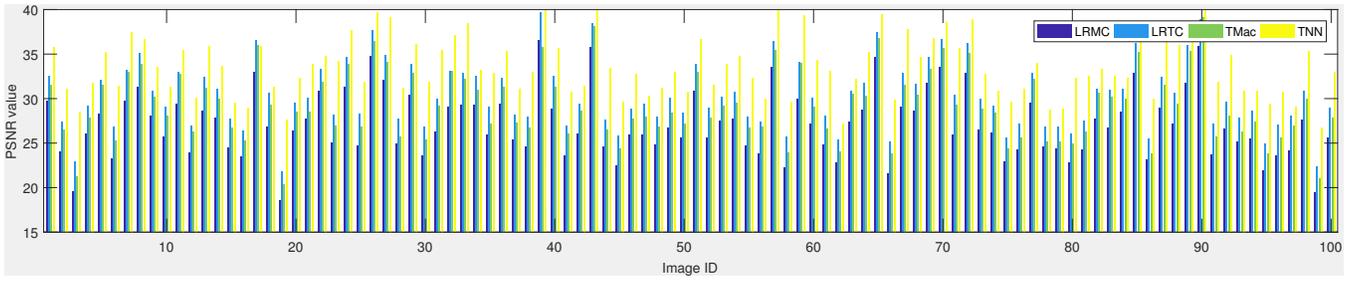

Figure 7: Comparison of the PSNR values obtained by using LRMC, LRTC, TMac and TNN. The rate of observed entries is $p = 0.5$.

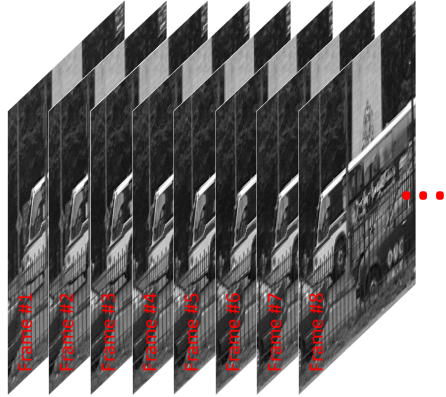

Figure 8: A grayscale video with $f$ sequences and frame size $h \times w$ can be formated as a tensor $\boldsymbol{\mathcal{M}} \in \mathbb{R}^{h \times f \times w}$.

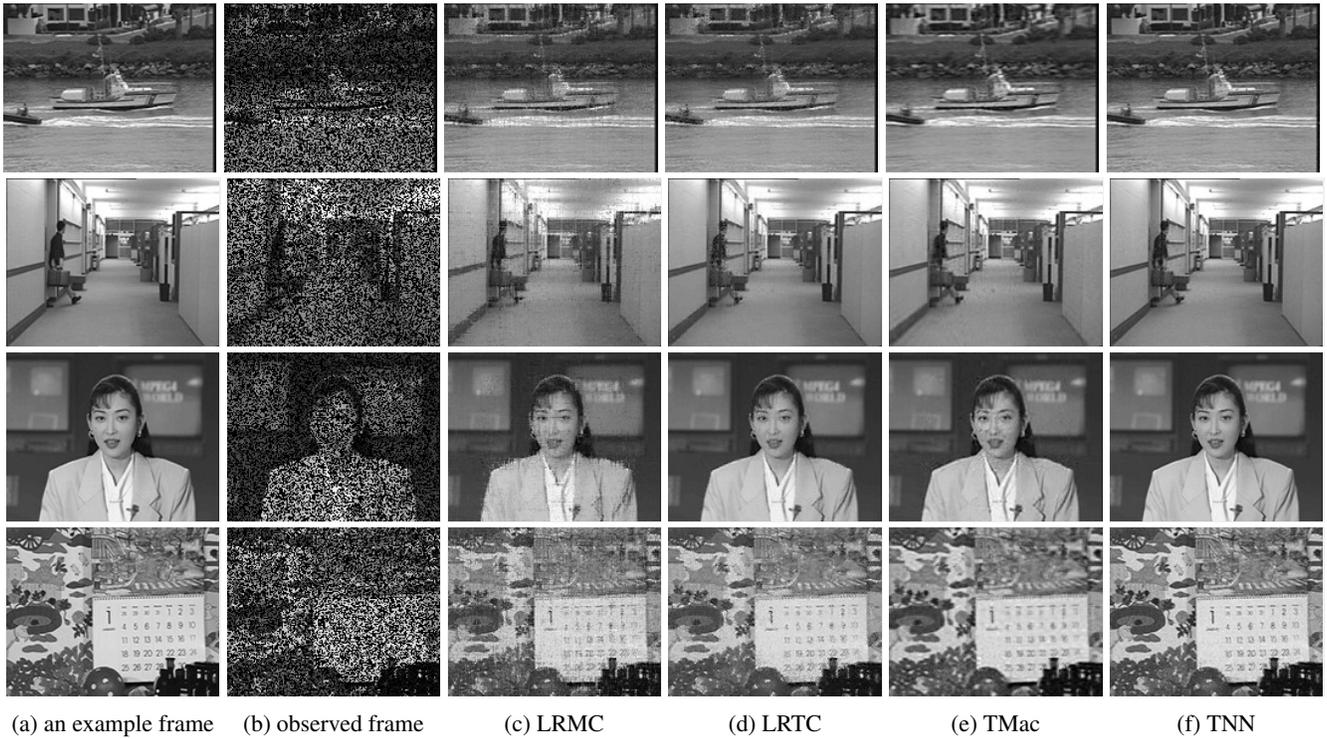

(a) an example frame   (b) observed frame   (c) LRMC   (d) LRTC   (e) TMac   (f) TNN

Figure 9: Examples for video recovery performance comparison. (a) Example frames from the sequences Coastguard, Hall, Akiyo and Mobile; (b) frames with partially observed entries (the rate is $p = 0.5$); (c)-(f) recovered frames by LRMC, LRTC, TMac and TNN, respectively.

Table 3: PSNR values of the compared methods.

| ID | Videos | LRMC | LRTC | TMac | TNN |
|---|---|---|---|---|---|
| 1 | Highway | 13.8 | 18.0 | 19.2 | **20.8** |
| 2 | Coastguard | 9.6 | 11.2 | 13.1 | **17.5** |
| 3 | Hall | 9.3 | 17.4 | 18.7 | **22.0** |
| 4 | Carphone | 10.9 | 16.7 | 18.3 | **20.3** |
| 5 | Bridge (close) | 10.5 | 17.8 | 17.6 | **20.9** |
| 6 | News | 8.6 | 15.4 | 16.7 | **20.3** |
| 7 | Grandma | 11.2 | 20.1 | 20.2 | **25.7** |
| 8 | Suzie | 14.5 | 17.4 | 19.9 | **19.7** |
| 9 | Miss America | 15.8 | 21.4 | 24.8 | **25.7** |
| 10 | Container | 8.4 | 17.8 | 17.3 | **29.0** |
| 11 | Foreman | 9.3 | 14.0 | 16.1 | **18.6** |
| 12 | Mother-daughter | 12.7 | 18.8 | 19.8 | **22.9** |
| 13 | Silent | 11.5 | 17.6 | 19.1 | **22.9** |
| 14 | Akiyo | 11.2 | 20.2 | 20.4 | **27.0** |
| 15 | Claire | 14.5 | 23.2 | 25.7 | **27.4** |

**Algorithm 2** Solve (13) by ADMM

**Input:** $A \in \mathbb{R}^{m \times (n_1 n_2 n_3)}, y \in \mathbb{R}^m$.
**Initialize:** $\mathcal{X}^0 = \mathcal{Z}^0 = \boldsymbol{\lambda}_2^0 = \mathbf{0}, \boldsymbol{\lambda}_1^0 = \mathbf{0}, \rho = 1.1, \mu_0 = 10^{-4}, \mu_{\max} = 10^{10}, \epsilon = 10^{-8}, k = 0$.
**while** not converged **do**

1. Update $\mathcal{X}^{k+1}$ by

$$\mathcal{X}^{k+1} = \underset{\mathcal{X}}{\operatorname{argmin}} \|\mathcal{X}\|_* + \frac{\mu_k}{2} \left\| \mathcal{X} - \mathcal{Z}^k + \frac{\boldsymbol{\lambda}_2^k}{\mu_k} \right\|_F^2 ;$$

2. Update $\mathcal{Z}^{k+1}$ by

$$\mathbf{z} = \underset{\mathcal{Z}}{\operatorname{argmin}} \, (A^\top A + I)^{-1} \left( -A^\top \frac{\boldsymbol{\lambda}_1^k}{\mu_k} + \frac{\operatorname{vec}(\boldsymbol{\lambda}_2^k)}{\mu_k} + A^\top \mathbf{y} + \operatorname{vec}(\mathcal{X}^{k+1}) \right) ;$$

$\mathcal{Z}^{k+1} \leftarrow \mathbf{z}$: reshape $\mathbf{z}$ to the tensor $\mathcal{Z}^{k+1}$ of size $n_1 \times n_2 \times n_3$.

3. Update the dual variables by

$$\boldsymbol{\lambda}_1^{k+1} = \boldsymbol{\lambda}_1^k + \mu_k (A \operatorname{vec}(\mathcal{Z}^{k+1}) - \mathbf{y});$$
$$\boldsymbol{\lambda}_2^{k+1} = \boldsymbol{\lambda}_2^k + \mu_k (\mathcal{X}^{k+1} - \mathcal{Z}^{k+1});$$

4. Update $\mu_{k+1}$ by $\mu_{k+1} = \min(\rho \mu_k, \mu_{\max})$;
5. Check the convergence conditions

$$\|\mathcal{X}^{k+1} - \mathcal{X}^k\|_\infty \leq \epsilon, \, \|\mathcal{Z}^{k+1} - \mathcal{Z}^k\|_\infty \leq \epsilon,$$
$$\|A \operatorname{vec}(\mathcal{Z}^{k+1}) - \mathbf{y}\|_\infty \leq \epsilon, \|\mathcal{X}^{k+1} - \mathcal{Z}^{k+1}\|_\infty \leq \epsilon.$$

**end while**

Without loss of generality, assume that $n_1 \leq n_2$. Then the complexity of the $\mathcal{X}$ update is $O(n_1 n_2 n_3 \log n_3 + n_1 n_2^2 n_3)$. For the update of $\mathcal{Z}$, beyond the pre-compute step of $(A^\top A + I)^{-1}$ which costs $(n_1 n_2 n_3)^2 m$, the per-iteration cost for the $\mathcal{Z}$ update is $O((n_1 n_2 n_3)^2 m)$.

Second, problem (8) can be reformulated (8) as follows

$$\min_{\mathcal{X}, \mathcal{E}} \|\mathcal{X}\|_*, \text{ s.t. } \mathcal{X} + \mathcal{E} = \mathcal{M}, \pi_\Omega(\mathcal{E}) = \mathbf{0}, \tag{14}$$

where $\pi_\Omega : \mathbb{R}^{n_1 \times n_2 \times n_3} \to \mathbb{R}^{n_1 \times n_2 \times n_3}$ is a linear operator that keeps the entries in $\Omega$ unchanged and sets those outside $\Omega$ (i.e., in $\Omega^c$) zeros. As $\mathcal{E}$ will compensate for the unknown entries of $\mathcal{X}$, the unknown entries of $\mathcal{X}$ are simply set as zeros. Then the

**Algorithm 3** Solve (14) by ADMM

**Input:** Observation samples $\mathcal{M}_{ijk}$, $(i, j, k) \in \Omega$, of tensor $\mathcal{M} \in \mathbb{R}^{n_1 \times n_2 \times n_3}$.
**Initialize:** $\mathcal{Y}_0 = \mathcal{E}_0 = 0$, $\rho = 1.1$, $\mu_0 = 10^{-4}$, $\mu_{\max} = 10^{10}$, $\epsilon = 10^{-8}$, $k = 0$.
**while** not converged **do**

1. Update $\mathcal{X}_{k+1}$ by

$$\mathcal{X}_{k+1} = \underset{\mathcal{X}}{\operatorname{argmin}} \|\mathcal{X}\|_* + \frac{\mu_k}{2} \left\| \mathcal{X} + \mathcal{E}_k - \mathcal{M} - \frac{\mathcal{Y}_k}{\mu_k} \right\|_F^2;$$

2. Update $\mathcal{E}_{k+1}$ by

$$\mathcal{E}_{k+1} = \pi_{\Omega^c}(\mathcal{M} - \mathcal{X}_{k+1} + \mathcal{Y}_k/\mu_k);$$

3. Update the dual variable by

$$\mathcal{Y}_{k+1} = \mathcal{Y}_k + \mu_k(\mathcal{M} - \mathcal{X}_{k+1} - \mathcal{E}_{k+1});$$

4. Update $\mu_{k+1}$ by $\mu_{k+1} = \min(\rho\mu_k, \mu_{\max})$;
5. Check the convergence conditions

$$\|\mathcal{X}_{k+1} - \mathcal{X}_k\|_\infty \leq \epsilon, \|\mathcal{E}_{k+1} - \mathcal{E}_k\|_\infty \leq \epsilon,$$
$$\|\mathcal{M} - \mathcal{X}_{k+1} - \mathcal{E}_{k+1}\|_\infty \leq \epsilon;$$

6. $k = k + 1$.

**end while**

partial augmented Lagrangian function of (14) is

$$L(\mathcal{X}, \mathcal{E}, \mathcal{Y}, \mu) = \|\mathcal{X}\|_* + \langle \mathcal{Y}, \mathcal{M} - \mathcal{X} - \mathcal{E} \rangle + \frac{\mu}{2} \|\mathcal{M} - \mathcal{X} - \mathcal{E}\|_F^2,$$

where $\mathcal{Y}$ is the dual variable and $\mu > 0$. Then we can update $\mathcal{X}$ and $\mathcal{E}$ alternately. See Algorithm 3 for the optimization details. The per-iteration complexity is $O\left(n_1 n_2 n_3 \log(n_3) + n_{(1)} n_{(2)}^2 n_3\right)$. Note that our solver is much more efficient that the one in [Zhang and Aeron, 2017] since we use the faster method to compute the proximal operator of TNN in [Lu *et al.*, 2018a].

## C Proof of Theorem 5

*Proof.* For $\mathcal{A} \in \mathbb{R}^{n_1 \times n_2 \times n_3}$, its degrees of freedom are the same as $\bar{\mathcal{A}}$ since the discrete Fourier transformation is the invertible. Assume that $\text{rank}_t(\mathcal{A}) = r$, then we have $\text{rank}(\bar{A}^{(i)}) \leq r$, $i = 1, \cdots, n_3$. Then $\bar{A}^{(i)}$ has at most $r(n_1 + n_2 - r)$ degrees of freedom, and thus $\bar{\mathcal{A}}$ has at most $r(n_1 + n_2 - r)n_3$ degrees of freedom. □

## D Proof of Theorem 3

In this section, we give the proof of Theorem 3. We first introduce some lemmas in subsection D.1. Then we give the complete proof of Theorem 3 in subsection D.2.

### D.1 Some Lemmas

**Lemma 7.** *(Subgradient of tensor nuclear norm) [Lu et al., 2018a] Let $\mathcal{A} \in \mathbb{R}^{n_1 \times n_2 \times n_3}$ with $\text{rank}_t(\mathcal{A}) = r$ and its skinny t-SVD be $\mathcal{A} = \mathcal{U} * \mathcal{S} * \mathcal{V}^*$. The subdifferential (the set of subgradients) of $\|\mathcal{A}\|_*$ is $\partial\|\mathcal{A}\|_* = \{\mathcal{U} * \mathcal{V}^* + \mathcal{W} | \mathcal{U}^* * \mathcal{W} = 0, \mathcal{W} * \mathcal{V} = 0, \|\mathcal{W}\| \leq 1\}$.*

**Lemma 8.** *Let $\mathcal{A}$ be an $n_1 \times n_2 \times n_3$ tensor whose entries are independent standard Gaussian random variables. Then, for any $\mathcal{U} \in \mathbb{R}^{n_1 \times k_1 \times n_3}$ with $\|\mathcal{U}\| \leq 1$ and $\mathcal{V} \in \mathbb{R}^{n_2 \times k_2 \times n_3}$ with $\|\mathcal{V}\| \leq 1$, we have*

$$\mathbb{E}\|\mathcal{U}^* * \mathcal{A} * \mathcal{V}\| \leq \sqrt{n_3}(\sqrt{k_1} + \sqrt{k_2}).$$

*Proof.* We denote $B$ as the set of block sparse vectors, i.e., $B^k = \left\{ \mathbf{x} \in \mathbb{R}^{kn_3} | \mathbf{x} = [\mathbf{x}_1^\top, \cdots, \mathbf{x}_i^\top \cdots, \mathbf{x}_{n_3}^\top] \right.$, with $\mathbf{x}_i \in \mathbb{R}^k$, and there exists $j$ such that $\mathbf{x}_j \neq 0$ and $\mathbf{x}_i = 0, i \neq j\}$. We also denote $S^k = \{\mathbf{x} \in \mathbb{R}^{kn_3} | \|\mathbf{x}\|_2 = 1\}$. Then, there exist

$\mathbf{p} \in B^{k_2} \cap S^{k_2}$ and $\mathbf{q} \in B^{k_1} \cap S^{k_1}$ such that

$$\|\mathcal{U}^* * \mathcal{A} * \mathcal{V}\|$$
$$= \|\bar{\mathbf{U}}^* \bar{\mathbf{A}} \bar{\mathbf{V}}\|$$
$$= \max_{\mathbf{p},\mathbf{q}} \langle \bar{\mathbf{U}}^* \bar{\mathbf{A}} \bar{\mathbf{V}} \mathbf{p}, \mathbf{q} \rangle$$
$$= \max_{\mathbf{p},\mathbf{q}} \langle (\boldsymbol{F}_{n_3} \otimes \boldsymbol{I}_{k_1}) \texttt{bcirc}(\mathcal{U}^* * \mathcal{A} * \mathcal{V})(\boldsymbol{F}_{n_3}^{-1} \otimes \boldsymbol{I}_{k_2}), \mathbf{q}\mathbf{p}^* \rangle$$
$$= \max_{\mathbf{p},\mathbf{q}} \langle \mathcal{A}, \mathcal{U} * \texttt{bcirc}^*((\boldsymbol{F}_{n_3}^{-1} \otimes \boldsymbol{I}_{k_1}) \mathbf{q}\mathbf{p}^* (\boldsymbol{F}_{n_3} \otimes \boldsymbol{I}_{k_2})) * \mathcal{V}^* \rangle,$$

where $\texttt{bcirc}^*$ is the joint operator of $\texttt{bcirc}$ which maps a matrix to a tensor. We denote

$$X_{\mathbf{p},\mathbf{q}} = \langle \mathcal{A}, \mathcal{U} * \texttt{bcirc}^*((\boldsymbol{F}_{n_3}^{-1} \otimes \boldsymbol{I}_{k_1}) \mathbf{q}\mathbf{p}^* (\boldsymbol{F}_{n_3} \otimes \boldsymbol{I}_{k_2})) * \mathcal{V}^* \rangle,$$

and it is a Gaussian variable. We also define

$$Y_{\mathbf{p},\mathbf{q}} = \sqrt{n_3}(\langle \mathbf{g}, \mathbf{p} \rangle + \langle \mathbf{h}, \mathbf{q} \rangle),$$

where $\mathbf{g} \in B^{k_2}, \mathbf{h} \in B^{k_1}$ and their entries in nonzero blocks are independent standard Gaussian random variables. Then, for $\mathbf{p}, \mathbf{p}_2 \in B^{k_2} \cap S^{k_2}$ and $\mathbf{q}, \mathbf{q}_2 \in B^{k_1} \cap S^{k_1}$, we have

$$\mathbb{E}\|X_{\mathbf{p},\mathbf{q}} - X_{\mathbf{p}_2,\mathbf{q}_2}\|_F^2$$
$$= \|\mathcal{U} * \texttt{bcirc}^*((\boldsymbol{F}_{n_3}^{-1} \otimes \boldsymbol{I}_{k_1})(\mathbf{q}\mathbf{p}^* - \mathbf{q}_2\mathbf{p}_2^*) \cdot$$
$$\quad (\boldsymbol{F}_{n_3} \otimes \boldsymbol{I}_{k_2})) * \mathcal{V}^*\|_F^2$$
$$\leq \|\mathcal{U}\|^2 \|\texttt{bcirc}^*((\boldsymbol{F}_{n_3}^{-1} \otimes \boldsymbol{I}_{k_1})(\mathbf{q}\mathbf{p}^* - \mathbf{q}_2\mathbf{p}_2^*) \cdot$$
$$\quad (\boldsymbol{F}_{n_3} \otimes \boldsymbol{I}_{k_2}))\|_F^2 \|\mathcal{V}\|^2$$
$$\leq n_3 \|\mathbf{q}\mathbf{p}^* - \mathbf{q}_2\mathbf{p}_2^*\|_F^2$$
$$\leq n_3 (\|\mathbf{p} - \mathbf{p}_2\|_2^2 + \|\mathbf{q} - \mathbf{q}_2\|_2^2)$$
$$\leq \mathbb{E}\|Y_{\mathbf{p},\mathbf{q}} - Y_{p_2,\mathbf{q}_2}\|_F^2.$$

Then, we have

$$\mathbb{E}\|\mathcal{U}^* * \mathcal{A} * \mathcal{V}\|$$
$$= \mathbb{E} \max_{\mathbf{p} \in B^{k_2} \cap S^{k_2}, \mathbf{q} \in B^{k_1} \cap S^{k_1}} X_{\mathbf{p},\mathbf{q}}$$
$$\leq \mathbb{E} \max_{\mathbf{p} \in B^{k_2} \cap S^{k_2}, \mathbf{q} \in B^{k_1} \cap S^{k_1}} Y_{\mathbf{p},\mathbf{q}} \tag{15}$$
$$= \sqrt{n_3}(\mathbb{E}\|\mathbf{g}\|_2 + \mathbb{E}\|\mathbf{h}\|_2)$$
$$= \sqrt{n_3}(\sqrt{k_2} + \sqrt{k_1}), \tag{16}$$

where (15) uses Corollary 3.14 in [Ledoux and Talagrand, 2013], and (16) is due to the facts that $\mathbf{g} \in B^{k_2}$ and $\mathbf{h} \in B^{k_1}$. The proof is completed. □

**Lemma 9.** *Let $\mathcal{A} \in \mathbb{R}^{n_1 \times n_2 \times n_3}$ be a randon tensor with i.i.d. Gaussian entries each with mean zero and variance one. Then, for any $\mathcal{U} \in \mathbb{R}^{n_1 \times k_1 \times n_3}$ with $\|\mathcal{U}\| \leq 1$ and $\mathcal{V} \in \mathbb{R}^{n_2 \times k_2 \times n_3}$ with $\|\mathcal{V}\| \leq 1$, we have*

$$\mathbb{P}\left[\|\mathcal{U}^* * \mathcal{A} * \mathcal{V}\| \geq \sqrt{n_3}(\sqrt{k_1} + \sqrt{k_2} + s)\right] \leq \exp(-s^2/2).$$

*Proof.* It is known that the matrix spectral norm is the 1-Lipschitz continuous, i.e., $|\|\boldsymbol{A}\| - \|\boldsymbol{B}\|| \leq \|\boldsymbol{A} - \boldsymbol{B}\|_F$. We show that $\|\mathcal{U}^* * \mathcal{A} * \mathcal{V}\|$ is $\sqrt{n_3}$-Lipschitz continuous. Indeed,

$$|\|\mathcal{U}^* * \mathcal{A} * \mathcal{V}\| - \|\mathcal{U}^* * \mathcal{B} * \mathcal{V}\||$$
$$= |\|\bar{\mathbf{U}}^* \bar{\mathbf{A}} \bar{\mathbf{V}}\| - \|\bar{\mathbf{U}}^* \bar{\mathbf{B}} \bar{\mathbf{V}}\||$$
$$\leq \|\bar{\mathbf{U}}^* \bar{\mathbf{A}} \bar{\mathbf{V}} - \bar{\mathbf{U}}^* \bar{\mathbf{B}} \bar{\mathbf{V}}\|_F$$
$$\leq \|\bar{\mathbf{U}}\| \|\bar{\mathbf{A}} - \bar{\mathbf{B}}\|_F \|\bar{\mathbf{V}}\|$$
$$\leq \|\bar{\mathbf{A}} - \bar{\mathbf{B}}\|_F = \sqrt{n_3} \|\mathcal{A} - \mathcal{B}\|_F.$$

Now the conclusion follows from the estimates on the expectation (Lemma 8) and Gaussian concentration (Proposition 5.34 in [Vershynin, 2010]). □

## D.2 Proof of Theorem 3

*Proof.* Denote $T$ by the set

$$T = \{\mathcal{U} * \mathcal{Y}^* + \mathcal{W} * \mathcal{V}^*, \mathcal{Y}, \mathcal{W} \in \mathbb{R}^{n \times r \times n_3}\},$$

and by $T^\perp$ its orthogonal complement. Then the projections onto $T$ and $T^\perp$ are respectively

$$\mathcal{P}_T(\mathcal{Z}) = \mathcal{U} * \mathcal{U}^* * \mathcal{Z} + \mathcal{Z} * \mathcal{V} * \mathcal{V}^* - \mathcal{U} * \mathcal{U}^* * \mathcal{Z} * \mathcal{V} * \mathcal{V}^*,$$

and

$$\mathcal{P}_{T^\perp}(\mathcal{Z}) = \mathcal{Z} - \mathcal{P}_T(\mathcal{Z}) = (\mathcal{I}_{n_1} - \mathcal{U} * \mathcal{U}^*) * \mathcal{Z} * (\mathcal{I}_{n_2} - \mathcal{V} * \mathcal{V}^*). \tag{17}$$

Let $\mathcal{M} = \mathcal{U} * \mathcal{S} * \mathcal{V}^*$, where $\mathcal{U} \in \mathbb{R}^{n_1 \times r \times n_3}$, $\mathcal{S} \in \mathbb{R}^{r \times r \times n_3}$ and $\mathcal{V} \in \mathbb{R}^{n_2 \times r \times n_3}$, be the skinny T-SVD of $\mathcal{M}$. Note that the normal cone of the tensor nuclear norm ball at $\mathcal{M}$ is given by the cone generated by the subdifferential at $\mathcal{M}$:

$$N_A(\mathcal{M})$$
$$= \text{cone}\{\mathcal{U} * \mathcal{V}^* + \mathcal{W} | \mathcal{U}^* * \mathcal{W} = 0, \mathcal{W} * \mathcal{V} = 0, \|\mathcal{W}\| \leq 1\}$$
$$= \{t\mathcal{U} * \mathcal{V}^* + \mathcal{W} | \mathcal{U}^* * \mathcal{W} = 0, \mathcal{W} * \mathcal{V} = 0, \|\mathcal{W}\| \leq t, t \geq 0\}.$$

Let $\mathcal{G}$ be a Gaussian random tensor with i.i.d. entries, each with mean zero and unit variance. Then the tensor

$$\mathcal{Z}(\mathcal{G}) = \|\mathcal{P}_{T^\perp}\mathcal{G}\|\mathcal{U} * \mathcal{V}^* + \mathcal{P}_{T^\perp}\mathcal{G},$$

is in the normal cone at $\mathcal{M}$. Here $\mathcal{P}_{T^\perp}$ is defined in (17). We then compute

$$\mathbb{E}\|\mathcal{G} - \mathcal{Z}(\mathcal{G})\|_F^2$$
$$= \mathbb{E}\|\mathcal{P}_T\mathcal{G} + \mathcal{P}_{T^\perp}\mathcal{G} - \mathcal{P}_T\mathcal{Z}(\mathcal{G}) - \mathcal{P}_{T^\perp}\mathcal{Z}(\mathcal{G})\|_F^2$$
$$= \mathbb{E}\|\mathcal{P}_T\mathcal{G} - \mathcal{P}_T\mathcal{Z}(\mathcal{G})\|_F^2$$
$$= \mathbb{E}\|\mathcal{P}_T\mathcal{G}\|_F^2 + \mathbb{E}\|\mathcal{P}_T\mathcal{Z}(\mathcal{G})\|_F^2 \tag{18}$$
$$= \mathbb{E}\|\mathcal{P}_T\mathcal{G}\|_F^2 + r\mathbb{E}\|\mathcal{P}_{T^\perp}\mathcal{G}\|^2, \tag{19}$$

where (18) follows because $\mathcal{P}_T\mathcal{G}$ and $\mathcal{P}_{T^\perp}\mathcal{G}$ are independent, and (19) uses the fact that $\|\mathcal{U} * \mathcal{V}^*\|_F = \sqrt{r}$.

Now, we consider to bound $\mathbb{E}\|\mathcal{P}_T\mathcal{G}\|_F^2$ and $\mathbb{E}\|\mathcal{P}_{T^\perp}\mathcal{G}\|^2$ in (19), respectively. First, we have

$$\mathbb{E}\|\mathcal{P}_T\mathcal{G}\|_F^2$$
$$= \mathbb{E}\langle \mathcal{P}_T\mathcal{G}, \mathcal{P}_T\mathcal{G}\rangle$$
$$= \mathbb{E}\langle \mathcal{P}_T\mathcal{G}, \mathcal{G}\rangle$$
$$= \mathbb{E}\langle \mathcal{U} * \mathcal{U}^* * \mathcal{G} + \mathcal{G} * \mathcal{V} * \mathcal{V}^* - \mathcal{U} * \mathcal{U}^* * \mathcal{G} * \mathcal{V} * \mathcal{V}^*, \mathcal{G}\rangle$$
$$= n_3 n_1 r + n_3 n_2 r - n_3 r^2, \tag{20}$$

where the last equation is obtained by direct computing on the definition of t-product.

Second, note that $\mathcal{P}_{T^\perp}\mathcal{G} = (\mathcal{I}_{n_1} - \mathcal{U} * \mathcal{U}^*) * \mathcal{G} * (\mathcal{I}_{n_2} - \mathcal{V} * \mathcal{V}^*)$. By Lemma 9, we have

$$\mathbb{P}[\|\mathcal{P}_{T^\perp}(\mathcal{G})\| \geq \sqrt{n_3}(\sqrt{n_1 - r} + \sqrt{n_2 - r} + s)] \leq \exp(-s^2/2).$$

Let $\mu_{T^\perp} = \sqrt{n_3}(\sqrt{n_1 - r} + \sqrt{n_2 - r})$. We have

$$\mathbb{E}[\|\mathcal{P}_{T^\perp}(\mathcal{G})\|^2]$$
$$= \int_0^\infty \mathbb{P}[\|\mathcal{P}_{T^\perp}(\mathcal{G})\|^2 > h]dh$$
$$\leq \mu_{T^\perp}^2 + \int_{\mu_{T^\perp}^2}^\infty \mathbb{P}[\|\mathcal{P}_{T^\perp}(\mathcal{G})\|^2 > h]dh$$
$$\leq \mu_{T^\perp}^2 + \int_0^\infty \mathbb{P}[\|\mathcal{P}_{T^\perp}(\mathcal{G})\|^2 > \mu_{T^\perp}^2 + t]dt$$
$$\leq \mu_{T^\perp}^2 + \int_0^\infty \mathbb{P}[\|\mathcal{P}_{T^\perp}(\mathcal{G})\| > \mu_{T^\perp} + \sqrt{t}]dt$$
$$\leq \mu_{T^\perp}^2 + \int_0^\infty \exp(-t/(2n_3))dt$$
$$= \mu_{T^\perp}^2 + 2n_3. \tag{21}$$

Combing (19), (20) and (21), we have

$$\mathbb{E}\left[\inf_{\mathcal{Z}\in N_A(\mathcal{M})}\|\mathcal{G}-\mathcal{Z}\|_F^2\right]$$
$$\leq n_3 r(n_1+n_2-r)+n_3 r((\sqrt{n_1-r}+\sqrt{n_2-r})^2+2)$$
$$\leq n_3 r(n_1+n_2-r)+n_3 r(2(n_1+n_2-2r)+2)$$
$$\leq 3n_3 r(n_1+n_2-r).$$

The proof is completed by using Proposition 3.6 in [Chandrasekaran *et al.*, 2012]. □

## E  Proof of Theorem 6

In this section, we give the proof of Theorem 6. We first introduce some lemmas in subsection E.1. Then we give the complete proof of Theorem 6 in subsection E.2.

We define the $\ell_{\infty,2}$-norm of the tensor $\mathcal{M}$ as

$$\|\mathcal{M}\|_{\infty,2}=\max\left\{\max_i\|\mathcal{M}(i,:,:)\|_F,\max_j\|\mathcal{M}(:,j,:)\|_F\right\}.$$

Define the projection $\mathcal{P}_{\Omega}(\mathcal{Z})=\sum_{ijk}\delta_{ijk}z_{ijk}\mathfrak{e}_{ijk}$, where $\delta_{ijk}=1_{(i,j,k)\in\Omega}$, where $1_{(\cdot)}$ is the indicator function. Also $\Omega^c$ denotes the complement of $\Omega$ and $\mathcal{P}_{\Omega^\perp}$ is the projection onto $\Omega^c$. Denote $T$ by the set

$$T=\{\mathcal{U}*\mathcal{Y}^*+\mathcal{W}*\mathcal{V}^*,\mathcal{Y},\mathcal{W}\in\mathbb{R}^{n\times r\times n_3}\},$$

and by $T^\perp$ its orthogonal complement. Then the projections onto $T$ and $T^\perp$ are respectively

$$\mathcal{P}_T(\mathcal{Z})=\mathcal{U}*\mathcal{U}^**\mathcal{Z}+\mathcal{Z}*\mathcal{V}*\mathcal{V}^*-\mathcal{U}*\mathcal{U}^**\mathcal{Z}*\mathcal{V}*\mathcal{V}^*,$$
$$\mathcal{P}_{T^\perp}(\mathcal{Z})=\mathcal{Z}-\mathcal{P}_T(\mathcal{Z})=(\mathcal{I}_{n_1}-\mathcal{U}*\mathcal{U}^*)*\mathcal{Z}*(\mathcal{I}_{n_2}-\mathcal{V}*\mathcal{V}^*),$$

For $i=1,\ldots,n_1, j=1,\ldots,n_2$ and $k=1,\ldots,n_3$, we define the random variable $\delta_{ijk}=1_{(i,j,k)\in\Omega}$, where $1_{(\cdot)}$ is the indicator function. Then the projection $\mathcal{R}_\Omega$ is given by

$$\mathcal{R}_\Omega(\mathcal{Z}):=\frac{1}{p}\mathcal{P}_\Omega(\mathcal{Z})=\sum_{i,j,k}\frac{1}{p}\delta_{ijk}z_{ijk}\mathfrak{e}_{ijk},$$

where $\mathfrak{e}_{ijk}=\mathring{\mathfrak{e}}_i*\dot{\mathfrak{e}}_k*\mathring{\mathfrak{e}}_j^*$ is an $n_1\times n_2\times n_3$ sized tensor with its $(i,j,k)$-th entry equaling 1 and the rest equaling 0. Also $\Omega^c$ denotes the complement of $\Omega$ and $\mathcal{P}_{\Omega^\perp}$ is the projection onto $\Omega^c$. By using (9)-(10), we have

$$\|\mathcal{P}_T(\mathfrak{e}_{ijk})\|_F^2\leq\frac{\mu r(n_1+n_2)}{n_1 n_2}=\frac{2\mu r}{n}, \text{ if } n_1=n_2=n. \tag{22}$$

### E.1  Some Lemmas

**Lemma 10.** *[Tropp, 2012] Consider a finite sequence $\{\mathbf{Z}_k\}$ of independent, random $n_1\times n_2$ matrices that satisfy the assumption $\mathbb{E}\mathbf{Z}_k=0$ and $\|\mathbf{Z}_k\|\leq R$ almost surely. Let*

$$\sigma^2=\max\{\|\sum_k\mathbb{E}[\mathbf{Z}_k\mathbf{Z}_k^*]\|,\max\{\|\sum_k\mathbb{E}[\mathbf{Z}_k^*\mathbf{Z}_k]\|\}.$$

*Then, for any $t\geq 0$, we have*

$$\mathbb{P}\left[\left\|\sum_k\mathbf{Z}_k\right\|\geq t\right]$$
$$\leq(n_1+n_2)\exp\left(-\frac{t^2}{2\sigma^2+\frac{2}{3}Rt}\right)$$
$$\leq(n_1+n_2)\exp\left(-\frac{3t^2}{8\sigma^2}\right), \text{ for } t\leq\frac{\sigma^2}{R}.$$

*Or, for any $c>0$, we have*

$$\left\|\sum_k\mathbf{Z}_k\right\|\geq 2\sqrt{c\sigma^2\log(n_1+n_2)}+cB\log(n_1+n_2), \tag{23}$$

*with probability at least $1-(n_1+n_2)^{1-c}$.*

**Lemma 11.** *Suppose $\Omega \sim Ber(p)$. Then with high probability,*

$$\|\mathcal{P}_T \mathcal{R}_\Omega \mathcal{P}_T - \mathcal{P}_T\| \leq \epsilon, \qquad (24)$$

*provided that $p \geq c_0 \epsilon^{-2}(\mu r \log(nn_3))/(nn_3)$ for some numerical constant $c_0 > 0$. For the tensor of rectangular frontal slices, we need $p \geq c_0 \epsilon^{-2}(\mu r \log(n_{(1)} n_3))/(n_{(2)} n_3)$.*

*Proof.* For any tensor $\mathcal{Z}$, we can write

$$(\mathcal{P}_T \mathcal{R}_\Omega \mathcal{P}_T - \mathcal{P}_T)(\mathcal{Z})$$
$$= \sum_{ijk} \left(p^{-1} \delta_{ijk} - 1\right) \langle \mathfrak{e}_{ijk}, \mathcal{P}_T(\mathcal{Z}) \rangle \mathcal{P}_T(\mathfrak{e}_{ijk})$$
$$:= \sum_{ijk} \mathcal{H}_{ijk}(\mathcal{Z})$$

where $\mathcal{H}_{ijk} : \mathbb{R}^{n \times n \times n_3} \to \mathbb{R}^{n \times n \times n_3}$ is a self-adjoint random operator with $\mathbb{E}[\mathcal{H}_{ijk}] = 0$. Define the matrix operator $\bar{\boldsymbol{H}}_{ijk} : \mathbb{B} \to \mathbb{B}$, where $\mathbb{B} = \{\bar{\boldsymbol{B}} : \mathcal{B} \in \mathbb{R}^{n \times n \times n_3}\}$ denotes the set consists of block diagonal matrices with the blocks as the frontal slices of $\bar{\mathcal{B}}$, as

$$\bar{\boldsymbol{H}}_{ijk}(\bar{\boldsymbol{Z}}) = \left(p^{-1} \delta_{ijk} - 1\right) \langle \mathfrak{e}_{ijk}, \mathcal{P}_T(\mathcal{Z}) \rangle \mathtt{bdiag}(\overline{\mathcal{P}_T(\mathfrak{e}_{ijk})}).$$

By the above definitions, we have $\|\mathcal{H}_{ijk}\| = \|\bar{\boldsymbol{H}}_{ijk}\|$ and $\|\sum_{ijk} \mathcal{H}_{ijk}\| = \|\sum_{ijk} \bar{\boldsymbol{H}}_{ijk}\|$. Also $\bar{\boldsymbol{H}}_{ijk}$ is self-adjoint and $\mathbb{E}[\bar{\boldsymbol{H}}_{ijk}] = 0$. To prove the result by the non-commutative Bernstein inequality, we need to bound $\|\bar{\boldsymbol{H}}_{ijk}\|$ and $\left\|\sum_{ijk} \mathbb{E}[\bar{\boldsymbol{H}}_{ijk}^2]\right\|$. First, we have

$$\|\bar{\boldsymbol{H}}_{ijk}\|$$
$$= \sup_{\|\bar{\boldsymbol{Z}}\|_F = 1} \|\bar{\boldsymbol{H}}_{ijk}(\bar{\boldsymbol{Z}})\|_F$$
$$\leq \sup_{\|\bar{\boldsymbol{Z}}\|_F = 1} p^{-1} \|\mathcal{P}_T(\mathfrak{e}_{ijk})\|_F \|\mathtt{bdiag}(\overline{\mathcal{P}_T(\mathfrak{e}_{ijk})})\|_F \|\mathcal{Z}\|_F$$
$$= \sup_{\|\bar{\boldsymbol{Z}}\|_F = 1} p^{-1} \|\mathcal{P}_T(\mathfrak{e}_{ijk})\|_F^2 \|\bar{\boldsymbol{Z}}\|_F$$
$$\leq \frac{2\mu r}{nn_3 p},$$

where the last inequality uses (22). On the other hand, by direct computation, we have $\bar{\boldsymbol{H}}_{ijk}^2(\bar{\boldsymbol{Z}}) = (p^{-1} \delta_{ijk} - 1)^2 \langle \mathfrak{e}_{ijk}, \mathcal{P}_T(\mathcal{Z}) \rangle \langle \mathfrak{e}_{ijk}, \mathcal{P}_T(\mathfrak{e}_{ijk}) \rangle \mathtt{bdiag}(\overline{\mathcal{P}_T(\mathfrak{e}_{ijk})})$. Note that $\mathbb{E}[(p^{-1}\delta_{ijk} - 1)^2] \leq p^{-1}$. We have

$$\left\|\sum_{ijk} \mathbb{E}[\bar{\boldsymbol{H}}_{ijk}^2(\bar{\boldsymbol{Z}})]\right\|_F$$
$$\leq p^{-1} \left\|\sum_{ijk} \langle \mathfrak{e}_{ijk}, \mathcal{P}_T(\mathcal{Z}) \rangle \langle \mathfrak{e}_{ijk}, \mathcal{P}_T(\mathfrak{e}_{ijk}) \rangle \mathtt{bdiag}(\overline{\mathcal{P}_T(\mathfrak{e}_{ijk})})\right\|_F$$
$$\leq p^{-1} \sqrt{n_3} \|\mathcal{P}_T(\mathfrak{e}_{ijk})\|_F^2 \left\|\sum_{ijk} \langle \mathfrak{e}_{ijk}, \mathcal{P}_T(\mathcal{Z}) \rangle\right\|_F$$
$$= p^{-1} \sqrt{n_3} \|\mathcal{P}_T(\mathfrak{e}_{ijk})\|_F^2 \|\mathcal{P}_T(\mathcal{Z})\|_F$$
$$\leq p^{-1} \sqrt{n_3} \|\mathcal{P}_T(\mathfrak{e}_{ijk})\|_F^2 \|\mathcal{Z}\|_F$$
$$= p^{-1} \|\mathcal{P}_T(\mathfrak{e}_{ijk})\|_F^2 \|\bar{\boldsymbol{Z}}\|_F$$
$$\leq \frac{2\mu r}{nn_3 p} \|\bar{\boldsymbol{Z}}\|_F.$$

This implies $\left\|\sum_{ijk} \mathbb{E}[\bar{\boldsymbol{H}}_{ijk}^2]\right\| \leq \frac{2\mu r}{nn_3 p}$. Let $\epsilon \leq 1$. By Lemma 10, we have

$$\mathbb{P}\left[\|\mathcal{P}_T\mathcal{R}_\Omega\mathcal{P}_T - \mathcal{P}_T\| > \epsilon\right]$$

$$=\mathbb{P}\left[\left\|\sum_{ijk}\mathcal{H}_{ijk}\right\| > \epsilon\right]$$

$$=\mathbb{P}\left[\left\|\sum_{ijk}\bar{\boldsymbol{H}}_{ijk}\right\| > \epsilon\right]$$

$$\leq 2nn_3 \exp\left(-\frac{3}{8} \cdot \frac{\epsilon^2}{2\mu r/(nn_3 p)}\right)$$

$$\leq 2(nn_3)^{1-\frac{3}{16}c_0},$$

where the last inequality uses $p \geq c_0 \epsilon^{-2} \mu r \log(nn_3)/(nn_3)$. Thus, $\|\mathcal{P}_T\mathcal{R}_\Omega\mathcal{P}_T - \mathcal{P}_T\| \leq \epsilon$ holds with high probability for some numerical constant $c_0$. □

**Lemma 12.** *Suppose that $\mathcal{Z}$ is fixed, and $\Omega \sim \text{Ber}(p)$. Then, with high probability,*

$$\|(\mathcal{R}_\Omega - \mathcal{I})\mathcal{Z}\| \leq c\left(\frac{\log(nn_3)}{p}\|\mathcal{Z}\|_\infty + \sqrt{\frac{\log(nn_3)}{p}}\|\mathcal{Z}\|_{\infty,2}\right),$$

*for some numerical constant $c > 0$.*

*Proof.* Denote the tensor $\mathcal{H}_{ijk} = \left(p^{-1}\delta_{ijk} - 1\right) z_{ijk}\mathfrak{e}_{ijk}$. Then we have

$$(\mathcal{R}_\Omega - \mathcal{I})\mathcal{Z} = \sum_{ijk}\mathcal{H}_{ijk}.$$

Note that $\delta_{ijk}$'s are independent random scalars. Thus, $\mathcal{H}_{ijk}$'s are independent random tensors and $\bar{\boldsymbol{H}}_{ijk}$'s are independent random matrices. Observe that $\mathbb{E}[\bar{\boldsymbol{H}}_{ijk}] = \boldsymbol{0}$ and $\|\bar{\boldsymbol{H}}_{ijk}\| \leq p^{-1}\|\mathcal{Z}\|_\infty$. We have

$$\left\|\sum_{ijk}\mathbb{E}[\bar{\boldsymbol{H}}_{ijk}^*\bar{\boldsymbol{H}}_{ijk}]\right\|$$

$$=\left\|\sum_{ijk}\mathbb{E}[\mathcal{H}_{ijk}^* * \mathcal{H}_{ijk}]\right\|$$

$$=\left\|\sum_{ijk}\mathbb{E}[(1-p^{-1}\delta_{ijk})^2]z_{ijk}^2(\mathring{\mathfrak{e}}_j * \mathring{\mathfrak{e}}_j^*)\right\|$$

$$=\left\|\frac{1-p}{p}\sum_{ijk}z_{ijk}^2(\mathring{\mathfrak{e}}_j * \mathring{\mathfrak{e}}_j^*)\right\|$$

$$\leq p^{-1}\max_j\left|\sum_{i,k}z_{ijk}^2\right|$$

$$\leq p^{-1}\|\mathcal{Z}\|_{\infty,2}^2.$$

A similar calculation yields $\left\|\sum_{ijk}\mathbb{E}[\bar{\boldsymbol{H}}_{ijk}^*\bar{\boldsymbol{H}}_{ijk}]\right\| \leq p^{-1}\|\mathcal{Z}\|_{\infty,2}^2$. Then the proof is completed by applying the matrix Bernstein inequality in (23). □

**Lemma 13.** *Suppose that $\mathcal{Z} \in \boldsymbol{T}$ is a fixed tensor and $\Omega \sim \text{Ber}(p)$. Then, with high probability,*

$$\|\mathcal{P}_T\mathcal{R}_\Omega(\mathcal{Z}) - \mathcal{Z}\|_{\infty,2} \leq \frac{1}{2}\sqrt{\frac{nn_3}{\mu r}}\|\mathcal{Z}\|_\infty + \frac{1}{2}\|\mathcal{Z}\|_{\infty,2},$$

*provided that $p \geq c_0\mu r\log(nn_3)/(nn_3)$.*

*Proof.* For fixed $\mathcal{Z} \in \boldsymbol{T}$ and fixed $b \in [n]$, the $b$-th column of the tensor $\mathcal{P}_{\boldsymbol{T}}\mathcal{R}_{\Omega}(\mathcal{Z}) - \mathcal{Z}$ can be written as

$$(\mathcal{P}_{\boldsymbol{T}}\mathcal{R}_{\Omega}(\mathcal{Z}) - \mathcal{Z}) * \mathring{\mathfrak{e}}_b$$
$$= \sum_{ijk} (p^{-1} - 1)\delta_{ijk} z_{ijk} \mathcal{P}_{\boldsymbol{T}}(\mathfrak{e}_{ijk}) * \mathring{\mathfrak{e}}_b$$
$$:= \sum_{ijk} \mathcal{H}_{ijk},$$

where $\mathcal{H}_{ijk}$'s are independent column tensors in $\mathbb{R}^{n \times 1 \times n_3}$ and $\mathbb{E}[\mathcal{H}_{ijk}] = \mathbf{0}$. Let $\mathbf{h}_{ijk} \in \mathbb{R}^{nn_3}$ be the column vector obtained by vectorizing $\mathcal{H}_{ijk}$. Then we have

$$\|\mathbf{h}_{ijk}\|$$
$$\leq p^{-1}|z_{ijk}| \|\mathcal{P}_{\boldsymbol{T}}(\mathfrak{e}_{ijk}) * \mathring{\mathfrak{e}}_b\|_F$$
$$\leq p^{-1}\|\mathcal{Z}\|_\infty \sqrt{\frac{2\mu r}{nn_3}}$$
$$\leq \frac{1}{c_0 \log(nn_3)} \sqrt{\frac{2nn_3}{\mu r}} \|Z\|_\infty.$$

We also have

$$\left| \sum_{ijk} \mathbb{E}\left[\mathbf{h}_{ijk}^* \mathbf{h}_{ijk}\right] \right|$$
$$= \left| \sum_{ijk} \mathbb{E}\left[\|\mathcal{H}_{ijk}\|_F^2\right] \right|$$
$$= \frac{1-p}{p} \sum_{ijk} z_{ijk}^2 \|\mathcal{P}_{\boldsymbol{T}}(\mathfrak{e}_{ijk}) * \mathring{\mathfrak{e}}_b\|_F^2.$$

Note that

$$\|\mathcal{P}_{\boldsymbol{T}}(\mathfrak{e}_{ijk}) * \mathring{\mathfrak{e}}_b\|_F^2$$
$$= \|\mathcal{U} * \mathcal{U}^* * \mathring{\mathfrak{e}}_i * \mathring{\mathfrak{e}}_k * \mathring{\mathfrak{e}}_j^* * \mathring{\mathfrak{e}}_b + (\mathcal{I} - \mathcal{U}*\mathcal{U}^*) * \mathring{\mathfrak{e}}_i * \mathring{\mathfrak{e}}_k * \mathring{\mathfrak{e}}_j^* * \mathcal{V} * \mathcal{V}^* * \mathring{\mathfrak{e}}_b\|_F$$
$$\leq \|\mathcal{U}*\mathcal{U}^* * \mathring{\mathfrak{e}}_i * \mathring{\mathfrak{e}}_k\|_F \|\mathring{\mathfrak{e}}_j^* * \mathring{\mathfrak{e}}_b\|_F + \|(\mathcal{I} - \mathcal{U}*\mathcal{U}^*) * \mathring{\mathfrak{e}}_i * \mathring{\mathfrak{e}}_k\| \|\mathring{\mathfrak{e}}_j^* * \mathcal{V} * \mathcal{V}^* * \mathring{\mathfrak{e}}_b\|_F$$
$$\leq \sqrt{\frac{\mu r}{nn_3}} \|\mathring{\mathfrak{e}}_j^* * \mathring{\mathfrak{e}}_b\|_F + \|\mathring{\mathfrak{e}}_j^* * \mathcal{V} * \mathcal{V}^* * \mathring{\mathfrak{e}}_b\|_F.$$

It follows that

$$\left| \sum_{ijk} \mathbb{E}[\mathbf{h}_{ijk}^* \mathbf{h}_{ijk}] \right|$$
$$= \frac{2}{p} \sum_{ijk} z_{ijk}^2 \frac{\mu r}{nn_3} \|\mathring{\mathfrak{e}}_j^* * \mathring{\mathfrak{e}}_b\|_F^2 + \frac{2}{p} \sum_{ijk} z_{ijk}^2 \|\mathring{\mathfrak{e}}_j^* * \mathcal{V} * \mathcal{V}^* * \mathring{\mathfrak{e}}_b\|_F^2$$
$$= \frac{2\mu r}{pnn_3} \sum_{ik} z_{ibk}^2 + \frac{2}{p} \sum_j \|\mathring{\mathfrak{e}}_j^* * \mathcal{V} * \mathcal{V}^* * \mathring{\mathfrak{e}}_b\|_F^2 \sum_{ik} z_{ijk}^2$$
$$\leq \frac{2\mu r}{pnn_3} \|\mathcal{Z}\|_{\infty,2}^2 + \frac{2}{p} \|\mathcal{V} * \mathcal{V}^* * \mathring{\mathfrak{e}}_b\|_F^2 \|\mathcal{Z}\|_{\infty,2}^2$$
$$\leq \frac{4\mu r}{pnn_3} \|\mathcal{Z}\|_{\infty,2}^2$$
$$\leq \frac{4}{c_0 \log(nn_3)} \|\mathcal{Z}\|_{\infty,2}^2.$$

We can bound $\|\sum_{ijk} \mathbb{E}[\mathbf{h}_{ijk}\mathbf{h}_{ijk}^*]\|$ by the same quantity in a similar manner. Treating $\mathbf{h}_{ijk}$'s as $nn_3 \times 1$ matrices and applying the matrix Bernstein inequality in (23) gives that w.h.p.

$$\|(\mathcal{P}_T \mathcal{R}_\Omega(\mathcal{Z}) - \mathcal{Z}) * \mathring{\mathfrak{e}}_b\|_F$$
$$= \left\|\sum_{ijk} \mathcal{H}_{ijk}\right\|_F$$
$$= \left\|\sum_{ijk} \mathbf{h}_{ijk}\right\|_F$$
$$\leq \frac{C}{c_0}\sqrt{\frac{2nn_3}{\mu r}}\|\mathcal{Z}\|_\infty + 4\sqrt{\frac{C}{c_0}}\|\mathcal{Z}\|_{\infty,2}$$
$$\leq \frac{1}{2}\sqrt{\frac{nn_3}{\mu r}}\|\mathcal{Z}\|_\infty + \frac{1}{2}\|\mathcal{Z}\|_{\infty,2},$$

provided that $c_0$ in the lemma statement is large enough. In a similar fashion, we prove that $\|\mathring{\mathfrak{e}}_a^* * (\mathcal{P}_T \mathcal{R}_\Omega(\mathcal{Z}) - \mathcal{Z})\|_F$ is bounded by the same quantity w.h.p. The lemma follows from a union bound over all $(a, b) \in [n] \times [n]$. □

**Lemma 14.** *Suppose that $\mathcal{Z} \in T$ is a fixed tensor and $\Omega \sim Ber(p)$. Then, with high probability,*

$$\|\mathcal{Z} - \mathcal{P}_T \mathcal{R}_\Omega(\mathcal{Z})\|_\infty \leq \epsilon \|\mathcal{Z}\|_\infty,$$

*provided that $p \geq c_0\epsilon^{-2}(\mu r \log(nn_3))/nn_3$ (for the tensor of rectangular frontal slice, $p \geq c_0 \epsilon^{-2}(\mu r \log(n_{(1)}n_3))/n_{(2)}$) for some numerical constant $c_0 > 0$.*

*Proof.* For any tensor $\mathcal{Z} \in T$, we write

$$\mathcal{P}_T \mathcal{R}_\Omega(\mathcal{Z}) = \sum_{ijk} p^{-1} \delta_{ijk} z_{ijk} \mathcal{P}_T(\mathfrak{e}_{ijk}).$$

The $(a, b, c)$-th entry of $\mathcal{P}_T \mathcal{R}_\Omega(\mathcal{Z}) - \mathcal{Z}$ can be written as a sum of independent random variables, i.e.,

$$\langle \mathcal{P}_T \mathcal{R}_\Omega(\mathcal{Z}) - \mathcal{Z}, \mathfrak{e}_{abc} \rangle$$
$$= \sum_{ijk}(p^{-1}\delta_{ijk} - 1)z_{ijk} \langle \mathcal{P}_T(\mathfrak{e}_{ijk}), \mathfrak{e}_{abc} \rangle$$
$$:= \sum_{ijk} t_{ijk},$$

where $t_{ijk}$'s are independent and $\mathbb{E}(t_{ijk}) = 0$. Now we bound $|t_{ijk}|$ and $|\sum_{ijk} \mathbb{E}[t_{ijk}^2]|$. First

$$|t_{ijk}|$$
$$\leq p^{-1}\|\mathcal{Z}\|_\infty \|\mathcal{P}_T(\mathfrak{e}_{ijk})\|_F \|\mathcal{P}_T(\mathfrak{e}_{abc})\|_F$$
$$\leq \frac{2\mu r}{nn_3 p}\|\mathcal{Z}\|_\infty.$$

Second, we have

$$\left|\sum_{ijk} \mathbb{E}[t_{ijk}^2]\right|$$
$$\leq p^{-1}\|\mathcal{Z}\|_\infty^2 \sum_{ijk} \langle \mathcal{P}_T(\mathfrak{e}_{ijk}), \mathfrak{e}_{abc} \rangle^2$$
$$= p^{-1}\|\mathcal{Z}\|_\infty^2 \sum_{ijk} \langle \mathfrak{e}_{ijk}, \mathcal{P}_T(\mathfrak{e}_{abc}) \rangle^2$$
$$= p^{-1}\|\mathcal{Z}\|_\infty^2 \|\mathcal{P}_T(\mathfrak{e}_{abc})\|_F^2$$
$$\leq \frac{2\mu r}{nn_3 p}\|\mathcal{Z}\|_\infty^2.$$

Let $\epsilon \leq 1$. By Lemma 10, we have

$$\mathbb{P}\left[|[\mathcal{P}_T \mathcal{R}_\Omega(\mathcal{Z}) - \mathcal{Z}]_{abc}| > \epsilon \|\mathcal{Z}\|_\infty\right]$$

$$= \mathbb{P}\left[\left|\sum_{ijk} t_{ijk}\right| > \epsilon \|\mathcal{Z}\|_\infty\right]$$

$$\leq 2 \exp\left(-\frac{3}{8} \cdot \frac{\epsilon^2 \|\mathcal{Z}\|_\infty^2}{2\mu r \|\mathcal{Z}\|_\infty^2/(nn_3 p)}\right)$$

$$\leq 2(nn_3)^{-\frac{3}{16}c_0},$$

where the last inequality uses $p \geq c_0 \epsilon^{-2} \mu r \log(nn_3)/(nn_3)$. Thus, $\|\mathcal{P}_T \mathcal{R}_\Omega(\mathcal{Z}) - \mathcal{Z}\|_\infty \leq \epsilon \|\mathcal{Z}\|_\infty$ holds with high probability for some numerical constant $c_0$. □

### E.2  Proof of Theorem 6

**Proposition 15.** *The tensor $\mathcal{M}$ is the unique optimal solution to (8) if the following conditions hold:*

1. $\|\mathcal{P}_T \mathcal{R}_\Omega \mathcal{P}_T - \mathcal{P}_T\| \leq \frac{1}{2}$.
2. *There exists a dual certificate $\mathcal{Y} \in \mathbb{R}^{n_1 \times n_2 \times n_3}$ which satisfies $\mathcal{P}_\Omega(\mathcal{Y}) = \mathcal{Y}$ and*

   (a) $\|\mathcal{P}_{T^\perp}(\mathcal{Y})\| \leq \frac{1}{2}$.
   
   (b) $\|\mathcal{P}_T(\mathcal{Y}) - \mathcal{U} * \mathcal{V}^\top\|_F \leq \frac{1}{4}\sqrt{\frac{p}{n_3}}$.

*Proof.* Consider any feasible solution $\mathcal{X}$ to (8) with $\mathcal{P}_\Omega(\mathcal{X}) = \mathcal{P}_\Omega(\mathcal{M})$. Let $\mathcal{G}$ be an $n \times n \times n_3$ tensor which satisfies $\|\mathcal{P}_{T^\perp}\mathcal{G}\| = 1$ and $\langle \mathcal{P}_{T^\perp}\mathcal{G}, \mathcal{P}_{T^\perp}(\mathcal{X} - \mathcal{M})\rangle = \|\mathcal{P}_{T^\perp}(\mathcal{X} - \mathcal{M})\|_*$. Such $\mathcal{G}$ always exists by duality between the tensor nuclear norm and the tensor spectral norm. Note that $\mathcal{U} * \mathcal{V}^* + \mathcal{P}_{T^\perp}\mathcal{G}$ is a subgradient of $\mathcal{Z}$ and $\mathcal{Z} = \mathcal{M}$, we have

$$\|\mathcal{X}\|_* - \|\mathcal{M}\|_* \geq \langle \mathcal{U} * \mathcal{V}^* + \mathcal{P}_{T^\perp}\mathcal{G}, \mathcal{X} - \mathcal{M}\rangle. \tag{25}$$

We also have $\langle \mathcal{Y}, \mathcal{X} - \mathcal{M}\rangle = \langle \mathcal{P}_\Omega \mathcal{Y}, \mathcal{P}_\Omega(\mathcal{X} - \mathcal{M})\rangle = 0$ since $\mathcal{P}_\Omega(\mathcal{Y}) = \mathcal{Y}$. It follows that

$$\|\mathcal{X}\|_* - \|\mathcal{M}\|_*$$
$$\geq \langle \mathcal{U} * \mathcal{V}^* + \mathcal{P}_{T^\perp}\mathcal{G} - \mathcal{Y}, \mathcal{X} - \mathcal{M}\rangle$$
$$= \|\mathcal{P}_{T^\perp}(\mathcal{X} - \mathcal{M})\|_* + \langle \mathcal{U} * \mathcal{V}^* - \mathcal{P}_T \mathcal{Y}, \mathcal{X} - \mathcal{M}\rangle - \langle \mathcal{P}_{T^\perp}\mathcal{Y}, \mathcal{X} - \mathcal{M}\rangle$$
$$\geq \|\mathcal{P}_{T^\perp}(\mathcal{X} - \mathcal{M})\|_* - \|\mathcal{U} * \mathcal{V}^* - \mathcal{P}_T \mathcal{Y}\|_F \|\mathcal{P}_T(\mathcal{X} - \mathcal{M})\|_F - \|\mathcal{P}_{T^\perp}\mathcal{Y}\| \|\mathcal{P}_{T^\perp}(\mathcal{X} - \mathcal{M})\|_*$$
$$\geq \frac{1}{2}\|\mathcal{P}_{T^\perp}(\mathcal{X} - \mathcal{M})\|_* - \frac{1}{4}\sqrt{\frac{p}{n_3}}\|\mathcal{P}_T(\mathcal{X} - \mathcal{M})\|_F,$$

where the last inequality uses the Conditions (1) and (2) in the proposition. Now, by using Lemma 16 below, we have

$$\|\mathcal{X}\|_* - \|\mathcal{M}\|_*$$
$$\geq \frac{1}{2}\|\mathcal{P}_{T^\perp}(\mathcal{X} - \mathcal{M})\|_* - \frac{1}{4}\sqrt{\frac{p}{n_3}} \cdot \sqrt{\frac{2n_3}{p}}\|\mathcal{P}_{T^\perp}(\mathcal{X} - \mathcal{M})\|_*$$
$$> \frac{1}{8}\|\mathcal{P}_{T^\perp}(\mathcal{X} - \mathcal{M})\|_*.$$

Note that the right hand side of the above inequality is strictly positive for all $\mathcal{X}$ with $\mathcal{P}_\Omega(\mathcal{X} - \mathcal{M}) = 0$ and $\mathcal{X} \neq \mathcal{M}$. Otherwise, we must have $\mathcal{P}_T(\mathcal{X} - \mathcal{M}) = \mathcal{X} - \mathcal{M}$ and $\mathcal{P}_T \mathcal{R}_\Omega \mathcal{P}_T(\mathcal{X} - \mathcal{M}) = 0$, contradicting the assumption $\|\mathcal{P}_T \mathcal{R}_\Omega \mathcal{P}_T - \mathcal{P}_T\| \leq \frac{1}{2}$. Therefore, $\mathcal{M}$ is the unique optimum. □

**Lemma 16.** *If $\|\mathcal{P}_T \mathcal{R}_\Omega \mathcal{P}_T - \mathcal{P}_T\| \leq \frac{1}{2}$, then we have*

$$\|\mathcal{P}_T \mathcal{Z}\|_F \leq \sqrt{\frac{2n_3}{p}} \|\mathcal{P}_{T^\perp}\mathcal{Z}\|_*, \ \forall \mathcal{Z} \in \{\mathcal{Z}' : \mathcal{P}_\Omega(\mathcal{Z}') = 0\}.$$

*Proof.* We deduce

$$\|\sqrt{p}\mathcal{R}_\Omega \mathcal{P}_T \mathcal{Z}\|_F$$
$$= \sqrt{\langle(\mathcal{P}_T \mathcal{R}_\Omega \mathcal{P}_T - \mathcal{P}_T)\mathcal{Z}, \mathcal{P}_T \mathcal{Z}\rangle + \langle \mathcal{P}_T \mathcal{Z}, \mathcal{P}_T \mathcal{Z}\rangle}$$
$$\geq \sqrt{\|\mathcal{P}_T \mathcal{Z}\|_F^2 - \|\mathcal{P}_T \mathcal{R}_\Omega \mathcal{P}_T - \mathcal{P}_T\| \|\mathcal{P}_T \mathcal{Z}\|_F^2}$$
$$\geq \frac{1}{\sqrt{2}}\|\mathcal{P}_T \mathcal{Z}\|_F, \tag{26}$$

where the last inequality uses $\|\mathcal{P}_T\mathcal{R}_\Omega\mathcal{P}_T - \mathcal{P}_T\| \leq \frac{1}{2}$. On the other hand, $\mathcal{P}_\Omega(\mathcal{Z}) = 0$ implies that $\mathcal{R}_\Omega(\mathcal{Z}) = 0$ and thus

$$\|\sqrt{p}\mathcal{R}_\Omega\mathcal{P}_T\mathcal{Z}\|_F = \|\sqrt{p}\mathcal{R}_\Omega\mathcal{P}_{T^\perp}\mathcal{Z}\|_F \leq \frac{1}{\sqrt{p}}\|\mathcal{P}_{T^\perp}\mathcal{Z}\|_F \leq \sqrt{\frac{n_3}{p}}\|\mathcal{P}_{T^\perp}\mathcal{Z}\|_*, \tag{27}$$

where the last inequality uses

$$\|\mathcal{A}\|_F = \frac{1}{\sqrt{n_3}}\|\bar{A}\|_F \leq \frac{1}{\sqrt{n_3}}\|\bar{A}\|_* \leq \sqrt{n_3}\|\mathcal{A}\|_*.$$

The proof is completed by combining (26) and (27). $\square$

Now we give the completed proof of Theorem 6.

*Proof (of Theorem 6).* First, as shown in Lemma 11, the Condition 1 of Proposition 15 holds with high probability. Now we construct a dual certificate $\mathcal{Y}$ which satisfies Condition 2 in Proposition 15. We do this using the Golfing Scheme [Gross, 2011]. For the choice of $p$ in Theorem 6, we have

$$p \geq \frac{c_0 \mu r (\log(nn_3))^2}{nn_3} \geq \frac{1}{nn_3}, \tag{28}$$

for some sufficiently large $c_0 > 0$. Set $t_0 := 20\log(nn_3)$. Assume that the set $\Omega$ of observed entries is generated from $\Omega = \cup_{t=1}^{t_0}\Omega_t$, where each $t$ and tensor index $(i,j,k)$, $\mathbb{P}[(i,j,k) \in \Omega_t] = q := 1-(1-p)^{1/t_0}$ and is independent of all others. Clearly this $\Omega$ has the same distribution as the original model. Let $\mathcal{W}_0 := \mathbf{0}$ and for $t = 1,\ldots,t_0$, define

$$\mathcal{W}_t = \mathcal{W}_{t-1} + \mathcal{R}_{\Omega_t}\mathcal{P}_T(\mathcal{U}*\mathcal{V}^* - \mathcal{P}_T\mathcal{W}_{t-1}),$$

where the operator $\mathcal{R}_{\Omega_t}$ is defined analogously to $\mathcal{R}_\Omega$ as $\mathcal{R}_{\Omega_t}(\mathcal{Z}) := \sum_{ijk} q^{-1} 1_{(i,j,k)\in\Omega_t} z_{ijk}\mathfrak{e}_{ijk}$. Then the dual certificate is given by $\mathcal{Y} := \mathcal{W}_{t_0}$. We have $\mathcal{P}_\Omega(\mathcal{Y}) = \mathcal{Y}$ by construction. To prove Theorem 6, we only need to show that $\mathcal{Y}$ satisfies Conditions 2 in Proposition 15 w.h.p.

**Validating Condition 2 (b).** Denote $\mathcal{D}_t := \mathcal{U}*\mathcal{V}^* - \mathcal{P}_T\mathcal{W}_k$ for $t = 0,\ldots,t_0$. By the definition of $\mathcal{W}_k$, we have $\mathcal{D}_0 = \mathcal{U}*\mathcal{V}^*$ and

$$\mathcal{D}_t = (\mathcal{P}_T - \mathcal{P}_T\mathcal{R}_{\Omega_t}\mathcal{P}_T)\mathcal{D}_{t-1}. \tag{29}$$

Obviously $\mathcal{D}_t \in T$ for all $t \geq 0$. Note that $\Omega_t$ is independent of $\mathcal{D}_{t-1}$ and by the choice of $p$ in Theorem 6, we have

$$q \geq \frac{p}{t_0} \geq \frac{c_0 \mu r \log(nn_3)}{nn_3}. \tag{30}$$

Applying Lemma 11 with $\Omega$ replaced by $\Omega_t$, we obtain that w.h.p.

$$\|\mathcal{D}_t\|_F \leq \|\mathcal{P}_T - \mathcal{P}_T\mathcal{R}_{\Omega_t}\mathcal{P}_T\|\|\mathcal{D}_{t-1}\|_F \leq \frac{1}{2}\|\mathcal{D}_{t-1}\|_F,$$

for each $t$. Applying the above inequality recursively with $t = t_0, t_0-1,\ldots,1$ gives

$$\|\mathcal{P}_T\mathcal{Y} - \mathcal{U}*\mathcal{V}^*\|_F = \|\mathcal{D}_{t_0}\|_F \leq \left(\frac{1}{2}\right)^{t_0}\|\mathcal{U}*\mathcal{V}^*\|_F$$

$$\leq \frac{1}{4nn_3}\cdot\sqrt{r} \leq \frac{1}{4\sqrt{nn_3}} \leq \frac{1}{4}\sqrt{\frac{p}{n_3}},$$

where the last inequality uses (28).

**Validating Condition 2 (a).** Note that $\mathcal{Y} = \sum_{t=1}^{t_0}\mathcal{R}_{\Omega_t}\mathcal{P}_T\mathcal{D}_{t-1}$ by construction. We have

$$\|\mathcal{P}_{T^\perp}\mathcal{Y}\|$$
$$\leq \sum_{t=1}^{t_0}\|\mathcal{P}_{T^\perp}(\mathcal{R}_{\Omega_t}\mathcal{P}_T - \mathcal{P}_T)\mathcal{D}_{t-1}\|$$
$$\leq \sum_{t=1}^{t_0}\|(\mathcal{R}_{\Omega_t} - \mathcal{I})\mathcal{P}_T\mathcal{D}_{t-1}\|.$$

Applying Lemma 12 with $\mathbf{\Omega}$ replaced by $\mathbf{\Omega}_t$ to the above inequality, we get that w.h.p.

$$\|\mathcal{P}_{T^\perp}\mathcal{Y}\|$$
$$\leq c\sum_{t=1}^{t_0}\left(\frac{\log(nn_3)}{q}\|\mathcal{D}_{t-1}\|_\infty + \sqrt{\frac{\log(nn_3)}{q}}\|\mathcal{D}_{t-1}\|_{\infty,2}\right)$$
$$\leq \frac{c}{\sqrt{c_0}}\sum_{t=1}^{t_0}\left(\frac{nn_3}{\mu r}\|\mathcal{D}_{t-1}\|_\infty + \sqrt{\frac{nn_3}{\mu r}}\|\mathcal{D}_{t-1}\|_{\infty,2}\right), \quad (31)$$

where the last inequality uses (30). Now we bound $\|\mathcal{D}_{t-1}\|_\infty$ and $\|\mathcal{D}_{t-1}\|_{\infty,2}$. Using (29) and repeatedly applying Lemma 14 with $\mathbf{\Omega}$ replaced as $\mathbf{\Omega}_t$, we obtain that w.h.p.

$$\|\mathcal{D}_{t-1}\|_\infty$$
$$=\|(\mathcal{P}_T - \mathcal{P}_T\mathcal{R}_{\Omega_{t-1}}\mathcal{P}_T)\cdots(\mathcal{P}_T - \mathcal{P}_T\mathcal{R}_{\Omega_1}\mathcal{P}_T)\mathcal{D}_0\|_\infty$$
$$\leq \left(\frac{1}{2}\right)^{t-1}\|\mathcal{U}*\mathcal{V}^*\|_\infty.$$

By Lemma 13 with $\mathbf{\Omega}$ replaced by $\mathbf{\Omega}_t$, we obtain that w.h.p.

$$\|\mathcal{D}_{t-1}\|_{\infty,2}$$
$$=\|(\mathcal{P}_T - \mathcal{P}_T\mathcal{R}_{\Omega_{t-1}}\mathcal{P}_T)\mathcal{D}_{t-2}\|_{\infty,2}$$
$$\leq \frac{1}{2}\sqrt{\frac{nn_3}{\mu r}}\|\mathcal{D}_{t-2}\|_\infty + \frac{1}{2}\|\mathcal{D}_{t-2}\|_{\infty,2}.$$

Using (29) and combining the last two display equations gives w.h.p.

$$\|\mathcal{D}_{t-1}\|_{\infty,2} \leq t\left(\frac{1}{2}\right)^{t-1}\sqrt{\frac{nn_3}{\mu r}}\|\mathcal{U}*\mathcal{V}^*\|_\infty + \left(\frac{1}{2}\right)^{t-1}\|\mathcal{U}*\mathcal{V}^*\|_{\infty,2}.$$

Substituting back to (31), we get w.h.p.

$$\|\mathcal{P}_{T^\perp}\mathcal{Y}\|$$
$$\leq \frac{c}{\sqrt{c_0}}\frac{nn_3}{\mu r}\|\mathcal{U}*\mathcal{V}^*\|_\infty \sum_{t=1}^{t_0}(t+1)\left(\frac{1}{2}\right)^{t-1} + \frac{c}{\sqrt{c_0}}\sqrt{\frac{nn_3}{\mu r}}\|\mathcal{U}*\mathcal{V}^*\|_{\infty,2}\sum_{t=1}^{t_0}\left(\frac{1}{2}\right)^{t-1}$$
$$\leq \frac{6c}{\sqrt{c_0}}\frac{nn_3}{\mu r}\|\mathcal{U}*\mathcal{V}^*\|_\infty + \frac{2c}{\sqrt{c_0}}\sqrt{\frac{nn_3}{\mu r}}\|\mathcal{U}*\mathcal{V}^*\|_{\infty,2}.$$

Now we proceed to bound $\|\mathcal{U}*\mathcal{V}^*\|_\infty$ and $\|\mathcal{U}*\mathcal{V}^*\|_{\infty,2}$. First, by the definition of t-product, we have

$$\|\mathcal{U}*\mathcal{V}^*\|_\infty$$
$$=\max_{ij}\left\|\sum_{t=1}^r \mathcal{U}(i,t,:)*\mathcal{V}(j,t,:)\right\|_\infty$$
$$\leq \max_{ij}\sum_{t=1}^r \|\mathcal{U}(i,t,:)\|_F\|\mathcal{V}(j,t,:)\|_F$$
$$\leq \max_{ij}\sum_{t=1}^r \frac{1}{2}\left(\|\mathcal{U}(i,t,:)\|_F^2 + \|\mathcal{V}(j,t,:)\|_F^2\right)$$
$$=\max_{ij}\frac{1}{2}\left(\|\mathcal{U}^**\mathring{\mathfrak{e}}_i\|_F^2 + \|\mathcal{V}^**\mathring{\mathfrak{e}}_j\|_F^2\right)$$
$$\leq \frac{\mu r}{nn_3},$$

Also, we have

$$\|\mathcal{U}*\mathcal{V}^*\|_{\infty,2} \leq \max\left\{\max_i\|\mathring{\mathfrak{e}}_i^**\mathcal{U}*\mathcal{V}^*\|_F, \max_j\|\mathcal{U}*\mathcal{V}^**\mathring{\mathfrak{e}}_j\|_F\right\} \leq \sqrt{\frac{\mu r}{nn_3}}.$$

It follows that w.h.p.

$$\|\mathcal{P}_{T^\perp}\mathcal{Y}\| \leq \frac{6c}{\sqrt{c_0}} + \frac{2c}{\sqrt{c_0}} \leq \frac{1}{2},$$

provided that $c_0$ is sufficiently large. This completes the proof of Theorem 6. $\square$